\documentclass[journal]{IEEEtran}
\usepackage{amsmath,amsfonts}
\usepackage{array}
\usepackage{algorithmic}
\usepackage{multirow}
\usepackage{graphicx} 
\usepackage{float} 
\usepackage[caption=false,font=footnotesize,labelfont=rm,textfont=rm]{subfig}
\usepackage{CJKutf8}
\usepackage{textcomp}
\usepackage{stfloats}
\usepackage{url}

\usepackage{cite}
\usepackage{algorithm}
\usepackage{bm}
\usepackage{booktabs}
\usepackage{tabularx}
\usepackage{multicol}
\usepackage{microtype}
\begin{document}
\begin{CJK}{UTF8}{gbsn}
\title{\textbf{Deeper Insights into Learning Performance of Stochastic Configuration Networks}}
\author{
Xiufeng Yan \\
  Artificial Intelligence Research Institute \\
  China University of Mining and Technology, Xuzhou, 22116, China\\
  Dianhui Wang*
  \thanks{\textit{\underline{Corresponding author}}: 
\textbf{dh.wang@deepscn.com}}\\
 Artificial Intelligence Research Institute \\
  China University of Mining and Technology, Xuzhou, 22116, China\\
    State Key Laboratory of Synthetical Automation for Process Industries \\
  Northeastern University, Shenyang 110819, China \\
  %% examples of more authors
  
}

\maketitle
\newtheorem{remark}{\bf Remark}

\begin{abstract}
Stochastic Configuration Networks (SCNs) are a class of randomized neural networks that integrate randomized algorithms within an incremental learning framework. A defining feature of SCNs is the supervisory mechanism, which adaptively adjusts the distribution to generate effective random basis functions, thereby enabling error-free learning. In this paper, we present a comprehensive analysis of the impact of the supervisory mechanism on the learning performance of SCNs. Our findings reveal that the current SCN framework evaluates the effectiveness of each random basis function in reducing residual errors using a lower bound on its error reduction potential, which constrains SCNs' overall learning efficiency. Specifically, SCNs may fail to consistently select the most effective random candidate as the new basis function during each training iteration. To overcome this problem, we propose a novel method for evaluating the hidden layer’s output matrix, supported by a new supervisory mechanism that accurately assesses the error reduction potential of random basis functions without requiring the computation of the Moore-Penrose inverse of the output matrix. This approach enhances the selection of basis functions, reducing computational complexity and improving the overall scalability and learning capabilities of SCNs. We introduce a Recursive Moore-Penrose Inverse-SCN (RMPI-SCN) training scheme based on the new supervisory mechanism and demonstrate its effectiveness through simulations over some benchmark datasets. Experiments show that RMPI-SCN outperforms the conventional SCN in terms of learning capability, underscoring its potential to advance the SCN framework for large-scale data modeling applications.
\end{abstract}


\sloppy
\section{Introduction}
 \IEEEPARstart{N}{eural} Networks (NNs) have achieved significant advancements in recent decades, primarily due to their universal approximation capabilities for modeling complex nonlinear mappings\cite{ref1,ref2,ref3,ref4} and their effectiveness in learning from data. Despite the strengths, traditional gradient-based training methods, such as Backpropagation (BP), face challenges including high computational costs, vulnerability to local minima, and sensitivity to hyperparameters. In contrast, randomized learning algorithms \cite{ref5,ref6,ref7,ref8,ref9,ref10} offer substantial advantages for constructing fast learner models with significantly lower computational costs. Randomized training typically follows a two-step paradigm: first, random basis functions are generated, with input weights and biases drawn from a specified distribution; second, the output weights are optimized using least squares. Igelink and Pao \cite{ref5} provided a theoretical justification for Random Vector Functional-links (RVFLs), demonstrating that when input weights and biases are randomly selected from a uniform distribution centered at zero and the output weights are optimized via least squares, RVFLs act as universal approximators in a probabilistic sense. The underlying idea is that if the target function can be represented as the integral of parametrized basis functions, then randomly sampling parameters from appropriate distributions can yield an accurate approximation. Recently, Needell et.al.\cite{ref11} revisited and extended this significant result, using a concentration inequality to bound the accuracy of Monte-Carlo integral approximations in non-asymptotic settings.
Compared with gradient-based training methods, randomized learning algorithms present a distinctive advantage: by employing least squares optimization, they avoid the computational complexity associated with iterative gradient-based searches, presenting a promising solution for building fast and efficient learner models. For practical implementations, RVFL requires two parameters: the number of hidden nodes and the radius of the uniform distribution's support. Tyukin and Prokhorov \cite{ref12} and Li and Wang \cite{ref13} have argued that these parameters should be data-dependent. Without a supervisory mechanism to adaptively adjust the distribution's support, the approximation capability of RVFL may be constrained, potentially resulting in limited learning and generalization performance. 

\noindent Stochastic Configuration Networks (SCNs), introduced by Wang and Li in \cite{ref14}, are a class of randomized learning models within an incremental learning framework. In each training iteration, input weights and biases are randomly selected from uniform distributions to generate candidate random basis functions. SCNs then evaluate the effectiveness of each random basis function in reducing the training residual error using the inequality $\| e_L \|^2 \leq r \| e_{L-1} \|^2$, where $r$ represents the learning rate and $L$ denotes the number of hidden nodes. If none of the candidates satisfy the inequality constraint, the support of the uniform distribution is expanded to include more suitable functions. This adaptive expansion enables SCNs to explore a broader function space, aligning the generated random basis functions more closely with the data.
Unlike Random Vector Functional-links (RVFL), where the configuration of random basis functions is entirely random, SCNs adaptively select random basis functions based on the training samples and current residuals, demonstrating greater flexibility and learning capacity. Due to their efficiency and effectiveness in large-scale data modeling, SCNs have attracted considerable research interest \cite{ref15,ref16,ref17,ref18,ref19,ref20,ref21,ref22,ref23,ref24,ref25}, encompassing both theoretical advancements and industrial applications. For instance, Wang and Dang \cite{ref15,ref16,ref17,ref18,ref19,ref20,ref21,ref22,ref23,ref24} proposed Recurrent Stochastic Configuration Networks (RSCN), which extended supervised randomized learning techniques to recurrent neural networks. Additionally, the 2D-SCN, introduced in \cite{ref27}, optimizes two-dimensional input representations to enhance performance in image data modeling. Similar to \cite{ref26}, the deep SCN proposed in \cite{ref19} aims to reduce reliance on Backpropagation (BP) for training deep neural networks. Successful industrial applications of SCNs have also been documented \cite{ref21,ref22,ref23,ref24,ref25,ref26}, underscoring their adaptability and efficiency in handling large-scale data in real-world settings.

\noindent An essential aspect of research on SCNs, and incremental learning with neural networks more broadly, is accurately evaluating the effectiveness of a new basis function in reducing residual error. This evaluation is pivotal, as it directly influences the selection of new basis functions and consequently dictates key factors such as training convergence, learning capacity, and model compactness. Despite extensive research efforts on the supervisory mechanism for enhancing the effectiveness and efficiency of SCNs \cite{ref16,ref17,ref18}, a fundamental issue remains: the estimate of the residual error $\| e_L \|$, used to evaluate a basis function $g_L$’s effectiveness in reducing the existing residual error $\| e_{L-1} \|$, is inconsistent with the actual residual error observed after $g_L$ is added. This discrepancy suggests that SCNs may not consistently select the most effective random basis functions, potentially resulting in slower convergence. Furthermore, a key advantage of SCNs over RVFL lies in the adaptability for adjusting uniform distributions, which directly depends on accurately assessing the effectiveness of random basis functions from the distribution. Consequently, imprecise evaluations might constrain SCNs’ overall learning capacity.

\noindent In this paper, we systematically investigate the supervisory mechanism of Stochastic Configuration Networks (SCNs) and analyze its effects on the learning performance of SCN-I and SCN-III, two primary SCN algorithms. The main contributions of this study are summarized as follows:

\begin{itemize}
  \item [1)] 
  We propose a new calculation method for $[H_{L-1}, h]^{\dagger} Y$ based on the recursive Moore-Penrose inverse, where $H_{L-1}$ is the output matrix of the current hidden layer with $L-1$ basis functions, $h$ is the output of a candidate basis function $g$, and $Y$ represents the training output. This method enables fast computation of the residual $Y - [H_{L-1}, h][H_{L-1}, h]^{\dagger} Y$ by avoiding the direct calculation of $[H_{L-1}, h]^{\dagger}$, thereby providing a rapid and accurate assessment of $g$’s effectiveness in reducing training residuals. This approach significantly enhances the selection of basis functions, as well as the scalability and efficiency of SCNs.       
  \item [2)]
  Building on this efficient calculation method, we establish necessary and sufficient conditions for the training residual errors $\{ e_L \}$ to satisfy $ \| e_L \| \leq \sqrt{r} \| e_{L-1} \|$ for SCN-III, ensuring that:
   \begin{equation}
   \label{eq1}
      \lim_{L \to \infty} \sup \frac{\| e_L \|}{\| e_{L-1} \|} \leq \sqrt{r},
   \end{equation}
   confirming that the sequence of training residual errors converges with order one at a rate of at most $\sqrt{r}$. 
  \item [3)]
  Based on this new inequality constraint, we propose the Recursive Moore-Penrose Inverse SCN (RMPI-SCN) training scheme. Additionally, we incorporate a hyperparameter $\alpha$ to control the increase of $r$ with network size, allowing dynamic adjustments in the learning rate to improve flexibility and convergence efficiency. Simulation results across various datasets show that RMPI-SCN outperforms the original SCNs in both learning and testing.
  
\end{itemize} 
 
\noindent The rest of the paper is organized as follows. Section 2 reviews the basic concepts of SCNs and discusses the differences between SCN-I and SCN-III. Section 3 presents the necessary and sufficient conditions for the training residual errors to satisfy $\| e_L \| \leq \sqrt{r} \| e_{L-1} \|$ in SCN-III, based on the recursive calculation of the Moore-Penrose inverse of the output matrix. This section also details the new supervisory mechanism and the RMPI-SCN. Section 4 exhibits the simulation results, and Section 5 concludes the paper.

\section{Related work}

\noindent Let $f: \mathbb{R}^n \rightarrow \mathbb{R}$ denote the target function, and let $f_{L-1}$ represent a Single Layer Feedforward Neural Network (SLFNN) comprising $L-1$ hidden nodes, expressed as follows:
\begin{equation}
    \label{eq2}
    \begin{array}{l}
    {f_{L-1}(x) = \sum_{i=1}^{L-1} \beta_i\ , }\\
   { g_i(x)= \sigma_i(w_i^T x + b_i), \quad 1 \leq i \leq L-1. }
    \end{array}
\end{equation}
Here, $\sigma_i$, $w_i$, $b_i$, and $\beta_i$ correspond to the activation function, input weight, bias, and output weight associated with the $i$-th basis function $g_i$, respectively. The primary objective of incremental learning is to establish criteria for selecting the next basis function $g_L$, such that the sequence of residual errors $\{e_L\}$, where $e_L = f - f_L$, converges to zero. The following result from Wang and Li \cite{ref14} demonstrates the universal approximation property of SCNs.

\noindent \textbf{Theorem 1} \cite{ref14}. Let $K$ be a compact subset of $\mathbb{R}^n$. Suppose $\Gamma$ is a collection of basis functions that are bounded in $L^2(K)$ and that $\text{span}(\Gamma)$ is dense in $L^2(K)$. Let $0 < r < 1$ and $\{u_L\}$ be a sequence of nonnegative real numbers that converges to 0, satisfying $u_L \leq 1 - r$ for each $L$. Define the residual as $e_{L-1} = f - f_{L-1}$, and let:
\begin{equation}
    \label{eq3}
    \delta_L = (1 - r - u_L) \|e_{L-1}\|^2.
\end{equation}
If the random basis function $g_L$ satisfies the inequality:
\begin{equation}
    \label{eq4}
    \langle e_{L-1}, g_L \rangle^2 \geq \|g_L\|^2 \delta_L,
\end{equation}
and the output weights are evaluated as:
\begin{equation}
    \label{eq5}
    \beta_L = \frac{\langle e_{L-1}, g_L \rangle}{\|g_L\|^2}, 
\end{equation}
then, we have $\lim_{L \to \infty} \|e_L\| = 0$, where $f_L = \sum_{i=1}^L \beta_i g_i$ and $\| \cdot \|$ denotes the $L^2$ norm.

\noindent Note that Theorem 1 ensures the universal approximation property of SCN at the algorithmic level, distinguishing it from the mathematical formulation of RVFL in \cite{ref5}. The density of $\text{span}(\Gamma)$ guarantees the existence of basis function $g_L$ satisfying the inequality in Eq. (\ref{eq4}) for some $r$ close to 1. The inequality constraint in Eq. (\ref{eq4}) serves two fundamental roles. First, since $u_L \to 0$ as $L \to \infty$ and $r < 1$, there exists a threshold $L_0$ such that for all $L \geq L_0$, $r + u_L < r^* < 1$ holds. This condition ensures that
$\lim_{L \to \infty} \sup \frac{\|e_L\|}{\|e_{L-1}\|} \leq \sqrt{r^*},$
meaning that the residual errors converge at a rate of at least $\sqrt{r^*}$. Second, the inequality evaluates the capability of the current uniform distribution to generate random basis functions that meet the error reduction criterion $\|e_L\| \leq \sqrt{r} \|e_{L-1}\|$.

\noindent Building on Theorem 1, Wang and Li proposed two foundational algorithms for SCNs in \cite{ref14}: SCN-I and SCN-III. To support the in-depth analysis and comparison of these algorithms, Table \ref{tab1} provides an overview of the terminologies and notations.

\begin{table}[h!]
\caption{Notations and Definitions}
\label{tab1}
\centering
\renewcommand{\arraystretch}{1.3}
\begin{tabular}{|>{\centering\arraybackslash}m{1.1cm}|m{6.5cm}|}
\hline
\textbf{Notation} & \multicolumn{1}{c|}{\textbf{Definition}} \\
\hline
$X$ & Training inputs, where $X = [x_1, x_2, \dots, x_N]^T$ and $x_t \in \mathbb{R}^d$ for $1 \leq t \leq N$. \\
\hline
$Y$ & Training outputs, where $Y = [y_1, y_2, \dots, y_N]^T$ and $y_t \in \mathbb{R}$ for $1 \leq t \leq N$. \\
\hline
$\sigma$ & Activation function applied to each basis function. \\
\hline
$g_i$ & Basis function, defined as $g_i(x) = \sigma_i(w_i^T x + b_i)$ for $x \in \mathbb{R}^d$. \\
\hline
$w_i$ & Input weight for $i$-th basis function $g_i$, where $w_i = [w_{i,1}, \dots, w_{i,d}]^T$ for $1 \leq i \leq L-1$. \\
\hline
$b_i$ & Bias for the $i$-th basis function $g_i$. \\
\hline
$\beta_i$ & Output weight of $g_i$. \\
\hline
$h_i$ & Output vector of $g_i$ on $X$, where $h_i = g_i(X)$. \\
\hline
$H_{L-1}$ & Output matrix of each basis function, where $H_{L-1} = [h_1, \dots, h_{L-1}]$. \\
\hline
$f_{L-1}$ & The current SCN, where $f_{L-1} = \sum_{i=1}^{L-1} \beta_i g_i$. \\
\hline
$e_{L-1}$ & The current training residual, where $e_{L-1} = Y - \sum_{i=1}^{L-1} \beta_i h_i$. \\
\hline
\end{tabular}
\end{table}

\begin{figure}[!h]
    \centering
    \includegraphics[width=1\linewidth]{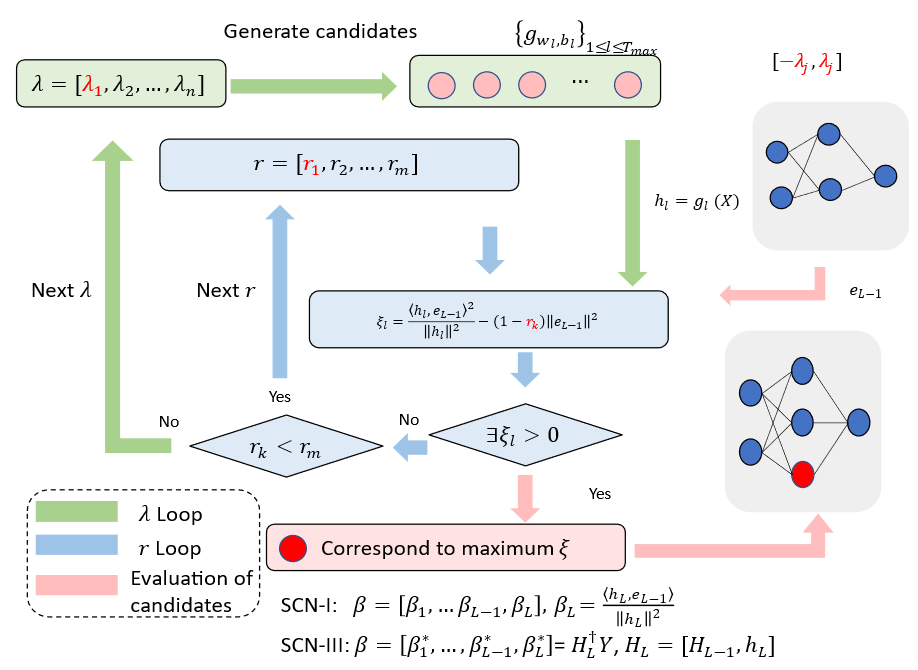}
    \caption{Training diagram of SCN}
    \label{fig1}
\end{figure}

\noindent To add the $L$-th hidden node, SCNs generate a batch of basis functions $\{g_l\}_{l=1}^{T_{\max}}$ with input weights $\{w_l\}_{l=1}^{T_{\max}}$ and biases $\{b_l\}_{l=1}^{T_{\max}}$ randomly sampled from the uniform distribution over $[-\lambda, \lambda]^d$ and $[-\lambda, \lambda]$, respectively. The output of each basis function $g_l$ on the input matrix is denoted as $h_l$ for $1 \leq l \leq T_{\max}$.
SCNs then execute a searching loop, incrementing the value of $r$ towards one to determine if the inequality in Eq. (4) can be satisfied by some $l$, ensuring the convergence of training residuals. If the inequality cannot be satisfied for any value of $r$, the parameter $\lambda$ will be increased to expand the support of the uniform distribution, allowing for the generation of another batch of random basis functions. This adjustment promotes the diversity of generated random basis functions, thereby enhancing the learning capabilities. Figure \ref{fig1} illustrates the general training process of SCNs.

\noindent The key distinction between SCN-I and SCN-III lies in the method used to update the output weights after the addition of the $L$-th basis function $g_L$. SCN-III employs:
\begin{equation}
    \label{eq6}
    \beta = [\beta_1^*, \dots, \beta_L^*]^T = \arg \min_{\beta} \| Y - \sum_{i=1}^L \beta_i h_i \| = H_L^{\dagger} Y,
\end{equation}
where $H_L = [H_{L-1}, h_L]$ is the augmented hidden layer output matrix, and $H_L^{\dagger}$ denotes the Moore-Penrose inverse of $H_L$. This approach allows for a comprehensive update of output weights, accounting for all basis functions, and consequently enhances convergence rates. In contrast, SCN-I updates only the output weight of the newly added basis function $g_L$ by:
\begin{equation}
    \label{eq7}
    \beta_L = \frac{\langle e_{L-1}, h_L \rangle}{\| h_L \|^2},
\end{equation}
and maintains the output weights of previous $L-1$ basis functions unchanged. This approach makes SCN-I more computationally efficient in resource-constrained environments. However, it is less adaptive and converges more slowly compared to SCN-III. Notably, the differences in the updates of output weights between SCN-I and SCN-III lead to distinct formulations of the residual error $e_L$. In SCN-I, the residual after adding $g_L$ is:
\begin{equation}
    \label{eq8}
    e_L^{\text{SCN-I}} = Y - \sum_{i=1}^{L-1} \beta_i h_i - \beta_L h_L = e_{L-1} - \beta_L h_L,
\end{equation}
where $\beta_L = \frac{\langle e_{L-1}, h_L \rangle}{\| h_L \|^2}$. In SCN-III, however, the residual error after the incorporation of $g_L$ is:
\begin{equation}
    \label{eq9}
    e_L^{\text{SCN-III}} = Y - \sum_{i=1}^L \beta_i^* h_i = Y - H_L H_L^{\dagger} Y.
\end{equation}
Both SCN-I and SCN-III evaluate the effectiveness of a basis function $g_l$ in reducing training residuals using:
\begin{equation}
    \label{eq10}
    \xi_l = \frac{\langle e_{L-1}, h_l \rangle^2}{\| h_l \|^2} - (1 - r - u_L) \| e_{L-1} \|^2.
\end{equation}
Since the term $(1 - r - u_L) \| e_{L-1} \|^2$ is independent of $h_l$, both algorithms select the new basis function from $\{g_l\}_{l=1}^{T_{\max}}$ by maximizing $\frac{\langle e_{L-1}, h_l \rangle^2}{\| h_l \|^2}$. It is worth noting that the objective function $\frac{\langle e_{L-1}, h \rangle^2}{\| h \|^2}$ is widely used in research that integrates optimization techniques within the framework of SCNs \cite{ref15}. However, the implication of $\frac{\langle e_{L-1}, h \rangle^2}{\| h \|^2}$ differs significantly for SCN-I and SCN-III due to the distinct ways in which the residual error $e_L$ is computed in each method.

\noindent In SCN-I, the random basis function $g_L$ with the maximum value of $\xi_L$ among $\{g_l\}_{l=1}^{T_{\max}}$ is indeed the optimal candidate for minimizing the training error. This is formally justified in Corollary 1, which shows that the inequality in Eq. \ref{eq4} is both a necessary and sufficient condition for $\| e_{L-1} - \beta_L h_L \| \leq \sqrt{r} \| e_{L-1} \|$ in SCN-I.

\noindent\textbf{Corollary 1.} Let $X=[x_1, x_2,\dots, x_N]^T$ and $Y=[y_1,y_2,\dots,y_N]^T$ denote the training inputs and outputs, respectively, where $x_t \in \mathbb{R}^d$ and $y_t \in \mathbb{R}$, $1 \leq t \leq N$. Given an activation function $\sigma$, the current training residual error is defined as:

\begin{equation}
     \label{eq11}
     e_{L-1} = Y - \sum_{i=1}^{L-1} \beta_i h_i,
\end{equation}
where $h_i = \sigma(Xw_i + b_i)$ for $1 \leq i \leq L-1$. Let $w_L$ and $b_L$ be the input weights and bias for the random basis function $g_L$, with $h_L = \sigma(Xw_L + b_L)$ as its output. Define:
\begin{equation}
    \label{eq12}
    \delta_L = (1 - r) \| e_{L-1} \|^2.
\end{equation}
Then, the condition:
\begin{equation}
    \label{eq13}
    \langle e_{L-1}, h_L \rangle^2 \geq \| h_L \|^2 \delta_L,
\end{equation}
holds if and only if:
\begin{equation}
    \label{eq14}
    \| e_{L-1} - \beta_L h_L \| \leq \sqrt{r} \| e_{L-1} \|,
\end{equation}
where $\beta_L$ is evaluated as
\begin{equation}
    \label{eq15}
    \beta_L = \frac{\langle e_{L-1}, h_L \rangle}{\| h_L \|^2}.
\end{equation}
In particular,
\begin{equation}
    \label{eq16}
    \begin{array}{cc}
    {\xi_L = \frac{\langle e_{L-1}, h_L \rangle^2}{\| h_L \|^2} - (1 - r) \| e_{L-1} \|^2}\\
    \hspace{-1.5em}{= r \| e_{L-1} \|^2 - \| e_L^{\text{SCN-I}}} \|^2.
    \end{array}
\end{equation}
\textbf{Proof.} Consider the expression
\begin{equation}
    \label{eq17}
    \begin{array}{cc}
        {\| e_{L-1} - \beta_L h_L \|^2 - r \| e_{L-1} \|^2}  \\
        {= \langle e_{L-1} - \beta_L h_L, e_{L-1} - \beta_L h_L \rangle - r \| e_{L-1} \|^2.} 
    \end{array}
\end{equation}
This can be rewritten as
\begin{equation}
    \label{eq18}
    \beta_L^2 \| h_L \|^2 - 2 \beta_L \langle e_{L-1}, h_L \rangle + (1 - r) \| e_{L-1} \|^2.
\end{equation}
The determinant of this quadratic equation in $\beta_L$ is given by
\begin{equation}
    \label{eq19}
    4 \langle e_{L-1}, h_L \rangle^2 - 4 \| h_L \|^2 \delta_L.
\end{equation}
The equation has a non-positive minimum if and only if
\begin{equation}
    \label{eq20}
    \langle e_{L-1}, h_L \rangle^2 \geq \| h_L \|^2 \delta_L.
\end{equation}
The minimum occurs at
\begin{equation}
    \label{eq21}
    \beta_L = \frac{\langle e_{L-1}, h_L \rangle}{\| h_L \|^2}.
\end{equation}
Substituting this value back into the quadratic equation yields the expression of $\xi_L$.

\noindent The proof of Corollary 1 demonstrates that $\| e_{L-1} \|^2 - \frac{\langle e_{L-1}, h_L \rangle^2}{\| h_L \|^2}$ represents the minimum value of the quadratic equation $\| e_{L-1} - \beta_L h_L \|^2$. Consequently, upon incorporating $g_L$ into the network, the residual error in SCN-I is updated as: $\| e_L^{(\text{SCN-I})} \|^2 = \| e_{L-1} \|^2 - \frac{\langle e_{L-1}, h_L \rangle^2}{\| h_L \|^2}$, demonstrating that the term $\frac{\langle e_{L-1}, h_L \rangle^2}{\| h_L \|^2}$ accurately quantifies the contribution of $g_L$ to the reduction of the training residuals.

\noindent In SCN-III, after the addition of the basis function $g_L$ with output $h_L$, the training error residual is $Y - H_L H_L^{\dagger} Y$, as presented in Eq. (\ref{eq9}). Notably, since $H_L^{\dagger} Y$ is the unique solution to the optimization problem $\arg \min_{\beta} \| Y - \sum_{i=1}^L \beta_i h_i \|$, we have $\| e_L^{(\text{SCN-III})} \|^2 \leq \| e_{L-1} - \beta_L h_L \|^2$. Hence,
\begin{equation}
    \label{eq22}
    \| e_L^{\text{SCN-III}} \|^2 \leq \| e_L^{\text{SCN-I}} \|^2.
\end{equation}
Combining the inequality in Eq. (22) with Eq. (16), we obtain:
\begin{equation}
    \label{eq23}
    \xi_L \leq r \| e_{L-1} \|^2 - \| e_L^{\text{SCN-III}} \|^2.
\end{equation}
The inequality in Eq. (\ref{eq23}) suggests that $\xi_L$ serves merely as a lower bound on the capability of the basis function $g_L$ to reduce training errors. Consequently, several concerns arise regarding the selection of basis functions based solely on $\xi$ in SCN-III.

\noindent First, it is unrealistic to assume that the basis function in $\{ g_l \}_{l=1}^{T_{\max}}$ with the largest value of $\xi$ will most effectively reduce training residuals. In fact, $\xi$ only reflects the basis function’s capability to reduce training error under the assumption that the output weights of the previous $L-1$ basis functions are unchanged. Consequently, using $\xi$ as a selection criterion in SCN-III may lead to slower convergence rates. Second, as $\xi$ serves as a lower bound for the capability of a basis function $g$ to reduce training residuals in SCN-III, evaluating basis functions by $\xi$ can become overly stringent as $L$ increases. As a result, SCN-III often selects $r$ that is extremely close to 1 to ensure the presence of random basis functions satisfying the inequality constraint. However, this approach may limit the model’s learning capacity by favoring random basis functions from a uniform distribution with a restricted support. Third, the value of $\xi$ is contingent upon the choice of $r$. Specifically, smaller values of $r$ impose stricter conditions on $h$ to satisfy $\| e_{L-1} - \beta_L h_L \| \leq \sqrt{r} \| e_{L-1} \|$. Given that $\| e_{L-1} - \beta_L h_L \|$ serves as a relatively loose upper bound for $\| Y - H_L H_L^{\dagger} Y \|$, a basis function that could significantly lower training errors may fail to meet the inequality constraint.
\section{The supervisory mechanism of SCN-III}

\noindent In this section, we systematically address the concerns regarding the supervisory mechanism of SCN-III in three steps. In subsection 3.1, we derive a necessary and sufficient condition for a basis function $g$ with output $h$ to satisfy $\| Y - H_L H_L^{\dagger} Y \| \leq \sqrt{r} \| e_{L-1} \|$ in SCN-III. The derivation is based on recursive calculations of the Moore-Penrose inverse.Building on these conditions, we propose a new supervisory mechanism for SCN-III in subsection 3.2, which further incorporates a hyperparameter $\alpha$ to control the increase of $r$ as $L \to \infty$. Subsection 3.3 details the algorithm of Recursive Moore-Penrose Inverse Stochastic Configuration Network (RMPI-SCN).

\subsection{The inequality constraint of SCN-III}

\noindent Consider a batch of basis functions $\{ g_l \}_{l=1}^{T_{\max}}$ with corresponding outputs on training samples $\{ h_l \}_{l=1}^{T_{\max}}$. Let $H_{L, l} = [H_{L-1}, h_l]$ for $1\leq l\leq T_{\max}$, where $H_{L-1}$ represents the hidden layer output matrix from the previous iteration. The main objective of SCN-III is to select $h_L$ by $\min_{1 \leq l \leq T_{\max}} \| H_{L, l} H_{L, l}^{\dagger} Y - Y \|$.Directly calculating $\| H_{L, l} H_{L, l}^{\dagger} Y - Y \|$ by computing $H_{L, l}^{\dagger}$ for each $l$ becomes computationally infeasible, particularly in large-scale data modeling. Nevertheless, since SCN is an incremental learning framework and $H_{L-1}^{\dagger}$ has been calculated in the previous iteration, recursive methods for calculating the Moore-Penrose inverse can significantly reduce the computational burden.

\noindent The following theorem presents a necessary and sufficient condition for $\| e_L \| \leq \sqrt{r} \| e_{L-1} \|$ in incremental learning with SLFNN. For brevity, we first present the case where the training output $Y = [y_1, y_2, \dots, y_N]^T$ with $y_t \in \mathbb{R}$ for $1 \leq t \leq N$. The extension to the general case where $y_t \in \mathbb{R}^m$ is straightforward and will be presented in Corollary 2.

\noindent \textbf{Theorem 2.} Suppose that $X$ and $Y$ are the training inputs and outputs respectively, where $X = [x_1, x_2, \dots, x_N]^T$, $Y = [y_1, y_2, \dots, y_N]^T$ and $x_t \in \mathbb{R}^d$, $y_t \in \mathbb{R}$, for $1 \leq i \leq N$. Given an activation function $\sigma$ and $0 < r < 1$, the current training error is $e_{L-1} = Y - H_{L-1} H_{L-1}^{\dagger} Y$ and $e_{L-1} \neq 0$, where $H_{L-1}^{\dagger}$ is the Moore-Penrose inverse of $H_{L-1}$, with $H_{L-1} = [h_1, h_2, \dots, h_{L-1}]$ and $h_i = \sigma(Xw_i + b_i)$, $1 \leq i \leq L-1$. Suppose that $h$ is the output of a candidate basis function $g$ on $X$. Define $H_L$, $\delta_L$, and $p_L$ as:
\begin{equation}
    \label{eq24}
    H_L = [H_{L-1}, h],
\end{equation}
\begin{equation}
    \label{eq25}
    \delta_L = \langle e_{L-1}, p_L \rangle^2 - (1 - r) \| p_L \|^2 \| e_{L-1} \|^2,
\end{equation}
\begin{equation}
    \label{eq26}
    p_L = h - H_{L-1} H_{L-1}^{\dagger} h.
\end{equation}
Then, $\| Y - H_L H_L^{\dagger} Y \| \leq \sqrt{r} \| e_{L-1} \|$ if and only if
\begin{equation}
    \label{eq27}
    p_L \neq 0,
\end{equation}
\begin{equation}
    \label{eq28}
    \frac{\langle e_{L-1}, p_L \rangle^2}{\| p_L \|^2} \geq (1 - r) \| e_{L-1} \|^2,
\end{equation}
\begin{equation}
    \label{eq29}
    \| \langle Y - e_{L-1}, p_L \rangle \| \leq \sqrt{\delta_L}.
\end{equation}
In particular, if $p_L \neq 0$, then
\begin{equation}
    \label{eq30}
    \| Y - H_L H_L^{\dagger} Y \| = \| e_{L-1} - \frac{\langle p_L, Y \rangle}{\| p_L \|^2} p_L \|.
\end{equation}
\textbf{Proof.} According to Greville \cite{ref27}, the Moore-Penrose inverse $H_L^{\dagger}$ of $H_L$ can be recursively calculated as follows:
\begin{equation}
    \label{eq31}
    d_L = H_{L-1}^{\dagger} h,
\end{equation}
 \begin{equation}
     \label{eq32}
     b_L = \begin{cases}
    (1 + d_L^T d_L)^{-1} d_L^T H_{L-1}^{\dagger}, & p_L = 0, \\
    p_L^{\dagger}, & p_L \neq 0,
    \end{cases} 
 \end{equation}
\begin{equation}
    \label{eq33}
    H_L^{\dagger} = \begin{bmatrix} H_{L-1}^{\dagger} - d_L b_L \\ b_L \end{bmatrix}.
\end{equation}
Thus, we have
\begin{equation}
    \label{eq34}
    \begin{array}{cc}
        {Y - H_L H_L^{\dagger} Y = Y - [H_{L-1}, h] \begin{bmatrix} H_{L-1}^{\dagger} - d_L b_L \\ b_L \end{bmatrix} Y} \\
    {= Y - [H_{L-1} H_{L-1}^{\dagger} - H_{L-1} d_L b_L + h b_L] Y}\\
    \hspace{-4.5em}{= e_{L-1} + H_{L-1} d_L b_L Y - h b_L Y} \\
    \hspace{-10.5em}{= e_{L-1} - p_L b_L Y.}\\
    \end{array}
\end{equation}
Define:
\begin{equation}
    \label{eq35}
    \tau_L = b_L Y = \begin{cases}
    (1 + d_L^T d_L)^{-1} d_L^T H_{L-1}^{\dagger} Y, & p_L = 0, \\
    p_L^{\dagger} Y, & p_L \neq 0.
\end{cases} 
\end{equation}
Since $\tau_L$ is a scalar, the inequality $\| H_L H_L^{\dagger} Y - Y \|^2 \leq r \| e_{L-1} \|^2$ can be written as
\begin{equation}
    \label{eq36}
    \| e_{L-1} - p_L \tau_L \|^2 \leq r \| e_{L-1} \|^2.
\end{equation}
If $p_L = 0$, the inequality holds if and only if $e_{L-1} = 0$. Thus, assume that $p_L \neq 0$ and $\tau_L = b_L Y = p_L^{\dagger} Y$. Since $p_L$ and $Y$ are two vectors in $\mathbb{R}^N$, we have
\begin{equation}
    \label{eq37}
    \tau_L = p_L^{\dagger} Y = \frac{\langle Y, p_L \rangle}{\| p_L \|^2}. 
\end{equation}
The inequality in Eq. (\ref{eq36}) becomes
\begin{equation}
    \label{eq38}
    \begin{array}{cc}
       {\| e_{L-1} - \tau_L p_L \|^2 - r \| e_{L-1} \|^2 }\\
       {= \langle e_{L-1} - \tau_L p_L, e_{L-1} - \tau_L p_L \rangle - r \| e_{L-1} \|^2 }\\
      \hspace{1.5em} { = \| p_L \|^2 \tau_L^2 - 2 \tau_L \langle e_{L-1}, p_L \rangle + (1 - r) \| e_{L-1} \|^2.}
    \end{array}
\end{equation}
This is a quadratic equation in $\tau_L$ and $\| p_L \|^2 > 0$, and it attains a non-positive value at $\tau_L = \frac{\langle Y, p_L \rangle}{\| p_L \|^2}$ if and only if its determinant is nonnegative and the value of $\tau_L$ lies between the two roots. The determinant of Eq. (\ref{eq38}) is $4 \langle e_{L-1}, p_L \rangle^2 - 4 (1 - r) \| p_L \|^2 \| e_{L-1} \|^2$. Therefore, $4 \langle e_{L-1}, p_L \rangle^2 - 4 (1 - r) \| p_L \|^2 \| e_{L-1} \|^2 \geq 0$ implies the inequality in Eq. (\ref{eq28}). For $\tau_L$ to lie between the two roots of the quadratic equation in Eq. (\ref{eq38}), the condition is $\| \tau_L - \frac{\langle e_{L-1}, p_L \rangle}{\| p_L \|^2} \| = \left\| \frac{\langle e_{L-1} - Y, p_L \rangle}{\| p_L \|^2} \right\| \leq \frac{\sqrt{\delta_L}}{\| p_L \|^2}$, which is equivalent to the condition in Eq. (\ref{eq29}). Finally, substituting $\tau_L = \frac{\langle Y, p_L \rangle}{\| p_L \|^2}$ into Eq. (\ref{eq34}) yields Eq. (\ref{eq30}), completing the proof.

\noindent In the proof of Theorem 2, the use of $\tau_L$ not only simplifies the deduction of the inequality constraint in Eq. (\ref{eq28})-(\ref{eq29}) but also enables the evaluation of $\| Y - H_L H_L^{\dagger} Y \|$ without computing $H_L^{\dagger}$ directly since $p_L = h - H_{L-1} H_{L-1}^{\dagger} h$ and $\tau_L = p_L^{\dagger} Y = \frac{\langle Y, p_L \rangle}{\| p_L \|^2}$ if $p_L \neq 0$. The calculation of $p_L$, $\tau_L$, and $\delta_L$ does not involve the possible calculation of $p_L^{\dagger}$. Hence, the calculation of Eq. (\ref{eq27})-(\ref{eq30}) only requires the Moore-Penrose inverse $H_{L-1}^{\dagger}$, which has been calculated after the addition of $g_{L-1}$ in SCN-III. Thus, the evaluation of $\min_{1 \leq l \leq T_{\max}} \| Y - H_{(L, l)} H_{(L, l)}^{\dagger} Y \|$ can be carried out without the calculation of $H_{(L, l)}^{\dagger}$ at all.

\noindent It is also important to note that the inequality constraints in Eq. (\ref{eq28}) and (\ref{eq29}) are the necessary and sufficient conditions for the inequality $\| Y - H_L H_L^{\dagger} Y \| \leq \sqrt{r} \| e_{L-1} \|$, which improves upon the previous constraint $\| e_{L-1} - \beta_L h_L \| \leq \sqrt{r} \| e_{L-1} \|$ (as in SCN-I). This ensures that the best candidate for reducing training residual errors is selected in each iteration of SCN-III. Compared with the older inequality constraint in Eq. (\ref{eq4}), the new conditions in Eq. (\ref{eq28})-(\ref{eq29}) will provide more candidate basis functions that satisfy the training error reduction constraint $\| e_L \| \leq \sqrt{r} \| e_{L-1} \|$, leading to a faster convergence rate. We present the following Corollary 2 for the general case $Y = [y_1, y_2, \dots, y_N]^T$ and $y_t \in \mathbb{R}^m$, $1 \leq t \leq N$.

\noindent \textbf{Corollary 2.} Suppose that $X$ and $Y$ are the training inputs and outputs respectively, where $X = [x_1, x_2, \dots, x_N]^T$, $Y = [y_1, y_2, \dots, y_N]^T$ and $x_t \in \mathbb{R}^d$, $y_t \in \mathbb{R}^m$, $1 \leq t \leq N$. Given an activation function $\sigma$ and $0 < r < 1$, let the current hidden layer output matrix be $H_{L-1} = [h_1, h_2, \dots, h_{L-1}]$ and $h_i = \sigma(Xw_i + b_i)$, $1 \leq i \leq L-1$, and $H_{L-1}^{\dagger}$ is the Moore-Penrose inverse of $H_{L-1}$. The current training error is defined as $e_{L-1} = Y - H_{L-1} H_{L-1}^{\dagger} Y = [e_{L-1,1}, \dots, e_{L-1,m}]$ where $e_{L-1,q} = Y_q - H_{L-1} H_{L-1}^{\dagger} Y_q$ and $Y_q = [y_{1,q}, y_{2,q}, \dots, y_{N,q}]^T$, $1 \leq q \leq m$. Let $h$ be the output of a candidate basis function $g$ on $X$ and define:
\begin{equation}
    \label{eq39}
    \begin{array}{cc}
         {\delta_{L,q} = \langle e_{L-1,q}, p_L \rangle^2 - (1 - r) \| p_L \|^2 \| e_{L-1,q} \|^2,}   \\
         1 \leq q \leq m, 
    \end{array}
\end{equation}
 \begin{equation}
     \label{eq40}
     p_L = h - H_{L-1} H_{L-1}^{\dagger} h.
 \end{equation}
Then, $\| Y_q - H_L H_L^{\dagger} Y_q \|^2 \leq r \| e_{L-1,q} \|^2$, $1 \leq q \leq m$, if and only if
\begin{equation}
    \label{eq41}
    p_L \neq 0,
\end{equation}
\begin{equation}
    \label{eq42}
    \frac{\langle e_{L-1,q}, p_L \rangle^2}{\| p_L \|^2} \geq (1 - r) \| e_{L-1,q} \|^2, \quad 1 \leq q \leq m,
\end{equation}
 \begin{equation}
     \label{eq43}
     \| \langle Y_q - e_{L-1,q}, p_L \rangle \| \leq \sqrt{\delta_{L,q}} ,\quad 1 \leq q \leq m.
 \end{equation}
In particular,
\begin{equation}
    \label{eq44}
    \| Y - H_L H_L^{\dagger} Y \|^2 = \sum_{q=1}^m \left\| e_{L-1,q} - \frac{\langle Y_q, p_L \rangle}{\| p_L \|^2} p_L \right\|^2.
\end{equation}
The proof is omitted as it follows directly from Theorem 2. Note that the conditions in Eq. (\ref{eq42})-(\ref{eq43}) are necessary and sufficient for $\| Y_q - H_L H_L^{\dagger} Y_q \|^2 \leq r \| e_{L-1,q} \|^2$, $1 \leq q \leq m$. However, these conditions are sufficient but not necessary for $\| Y - H_L H_L^{\dagger} Y \|^2 \leq r \| e_{L-1} \|^2$. If the training error reduction is evaluated using the inequality $\| Y - H_L H_L^{\dagger} Y \|^2 \leq r \| e_{L-1} \|^2$, Eq. (\ref{eq44}) offers a practical means for rapid evaluation.

\subsection{A supervisory mechanism with controlled learning rate r }

\noindent The parameter $r$ directly determines the evaluation of the candidates of basis functions in each iteration of the training of SCN through the inequality $\| e_L \| \leq \sqrt{r} \| e_{L-1} \|$. It not only ensures the rate of convergence but also determines whether the support of the uniform distribution should be expanded. The common approach is to iteratively search for $r$ within a sequence starting from $0.9$ to $1$. As the size of the network increases, finding a random basis function that satisfies the inequality $\| e_L \| \leq \sqrt{r} \| e_{L-1} \|$ becomes more difficult. To address this challenge, an upper bound $r_{\max}$ close to one is often set to ensure there are candidate functions available as the network grows. However, increasing $r$ does not inherently improve the capacity of random basis functions to reduce training errors; rather, it merely lowers the error reduction requirement. Fixing this requirement by imposing $\| e_L \| \leq \sqrt{r_{\max}} \| e_{L-1} \|$ at each iteration limits the SCN’s ability to adaptively explore optimal random weights, potentially leading to inefficient learning.

\noindent To resolve these issues, we propose a mechanism that gradually increases $r$ in accordance with network size, while ensuring convergence of training residual errors. This adaptive mechanism incorporates a hyperparameter $\alpha$, which modulates the rate of $r$’s increase based on the complexity of the data. Given $0 < r < 1$, $0 < \alpha$, if $\{ x_L \}$ is a sequence of non-negative real numbers such that $x_L \leq r^{(1 + 1/L)^{\alpha}} x_{L-1}$ for each $L$, then $\lim_{L \to \infty} x_L = 0$ evidently since
\begin{equation}
    \label{eq45}
    \lim_{L \to \infty} \sup \frac{x_L}{x_{L-1}} \leq \lim_{L \to \infty} r^{(1 + 1/L)^{\alpha}} = r^{\alpha}.
\end{equation}
Besides, $r^{(1 + 1/L)^{\alpha}}$ is a monotone increasing function with respect to $L$, and its supremum is $r^{\alpha}$. By controlling the rate at which $r$ increases using $\alpha$, we ensure that the SCN can dynamically adjust the range of random weights explored, balancing computational efficiency and model performance. The hyperparameter $\alpha$ thus provides a flexible means to control the speed of convergence, adapting to the complexity of the data being modeled.

\noindent Specifically, the logarithm of $r^{(1 + 1/L)^{\alpha}}$ is $\alpha (1 + 1/L) \ln r$ whose derivative with respect to $L$ is $-\frac{\alpha \ln r}{L^2}$. This means that a higher value of $\alpha$ causes $r$ to converge more rapidly to $r^{\alpha}$, allowing the model to relax the error reduction constraint at a faster rate as the network size grows. In practice, this provides a flexible way to adjust the learning rate based on the complexity of the data and the depth of the network, ensuring a more efficient search for suitable random basis functions as the model trains.

\begin{figure}[htpb]
    \centering
    \includegraphics[width=1\linewidth]{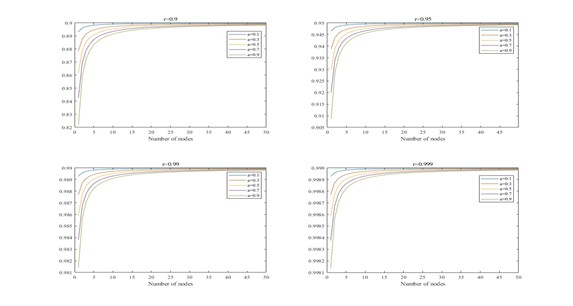}
    \caption{The curve of $r_L = r^{(1 + 1/L)^\alpha}$ for $\alpha = [0.1, 0.3, 0.5, 0.7, 0.9]$.}
    \label{fig2}
\end{figure}

 \noindent We now embed the design of $r^{(1 + 1/L)^\alpha}$ into the inequality constraints established in Corollary 2 to present the following Corollary 3.

\noindent \textbf{Corollary 3.} Suppose that $X$ and $Y$ are the training inputs and outputs respectively, where $X = [x_1, x_2, \dots, x_N]^T$, $Y = [y_1, y_2, \dots, y_N]^T$ and $x_t \in \mathbb{R}^d$, $y_t \in \mathbb{R}^m$, $1 \leq t \leq N$. Given a fixed activation function $\sigma$ and $0 < r < 1$, let the current hidden layer output matrix be $H_{L-1} = [h_1, \dots, h_{L-1}]$ and $h_i = \sigma(Xw_i + b_i)$, $1 \leq i \leq L-1$, and $H_{L-1}^{\dagger}$ is the Moore-Penrose inverse of $H_{L-1}$. The current training error is defined as $e_{L-1} = Y - H_{L-1} H_{L-1}^{\dagger} Y = [e_{L-1,1}, \dots, e_{L-1,m}]$, where $e_{L-1,q} = Y_q - H_{L-1} H_{L-1}^{\dagger} Y_q$ and $Y_q = [y_{1,q}, y_{2,q}, \dots, y_{N,q}]^T$, $1 \leq q \leq m$. Let $h$ be the output of a candidate basis function $g$ on $X$. Define $p_L$, $r_L$, and $\delta_{L,q}$ as:
\begin{equation}
    \label{eq46}
    p_L = h - H_{L-1} H_{L-1}^{\dagger} h,
\end{equation}
\begin{equation}
    \label{eq47}
    r_L = r^{(1 + 1/L)^\alpha},
\end{equation}
\begin{equation}
    \label{eq48}
    \begin{array}{cc}
         {\delta_{L,q} = \langle e_{L-1,q}, p_L \rangle^2 - (1 - r_L) \| p_L \|^2 \| e_{L-1,q} \|^2,}   \\
         { 1 \leq q \leq m.}
    \end{array}
\end{equation}
Then, $\| Y_q - H_L H_L^{\dagger} Y_q \|^2 \leq r_L \| e_{L-1,q} \|^2$, $1 \leq q \leq m$, if and only if
\begin{equation}
    \label{eq49}
    p_L \neq 0,
\end{equation}
\begin{equation}
    \label{eq50}
    \frac{\langle e_{L-1,q}, p_L \rangle^2}{\| p_L \|^2} \geq (1 - r_L) \| e_{L-1,q} \|^2, \quad 1 \leq q \leq m,
\end{equation}
\begin{equation}
    \label{eq51}
    \| \langle Y_q - e_{L-1,q}, p_L \rangle \| \leq \sqrt{\delta_{L,q}}, \quad 1 \leq q \leq m.
\end{equation}
In particular,
\begin{equation}
    \label{eq52}
    \| Y - H_L H_L^{\dagger} Y \|^2 = \sum_{q=1}^m \left\| e_{L-1,q} - \frac{\langle Y_q, p_L \rangle}{\| p_L \|^2} p_L \right\|^2.
\end{equation}

\subsection{Algorithm description}
\noindent This subsection details the algorithmic structure of the proposed Recursive Moore-Penrose Inverse SCN (RMPI-SCN). The general training diagram of SCN consists of two main components: configuration of random basis functions and evaluation of output weights. The proposed RMPI-SCN differs from SCN-III in the following respects.

\noindent First, RMPI-SCN cancels the inner iterative searching for $r$ and applies the design $r_L = r^{(1 + 1/L)^\alpha}$ in Corollary 3. This ensures adaptive learning while reducing computational overhead by systematically increasing $r$ as the network grows. Second, it adopts the constraints in Eq. (\ref{eq50})-(\ref{eq51}). These new constraints enhance the selection of random basis functions, providing a more efficient method for reducing training errors while maintaining theoretical guarantees for convergence. Third, it requires an initialization for the first basis function $g_1$ and the Moore-Penrose inverse $h_1^{\dagger}$ of its output $h_1$. This initialization is crucial for ensuring that the subsequent steps of RMPI-SCN build on a stable foundation, setting the stage for efficient training and accurate function approximation. We first present the pseudocode for the RMPI-SCN in the following and then detail the involved calculations.

\begin{algorithm}
\caption{Pseudocode for RMPI-SCN}
\begin{algorithmic}[1]
\STATE Given inputs $X = \{x_1, x_2, \dots, x_N\}$, $x_t \in \mathbb{R}^d$ and outputs $Y = \{y_1, y_2, \dots, y_N\}$, $y_t \in \mathbb{R}^m$. Set a maximum number of epochs $L_{\max}$, an error tolerance $\epsilon$, and the number of candidate nodes $T_{\max}$. Choose a set of increasing positive scalars $\Upsilon = \{\lambda_1, \dots, \lambda_K\}$, $0 < r < 1$ and $0 < \alpha < 1$.
\hrule
\STATE Initialize $X = [x_1, x_2, \dots, x_N]^T$, $Y = [y_1, y_2, \dots, y_N]^T$, $e_0 = Y$, and empty sets $W$ and $B$.
\STATE Randomly assign input weights $w_j$ and $b_j$, $1 \leq j \leq T_{\max}$ from $[-\lambda_1, \lambda_1]^d$ and $[-\lambda_1, \lambda_1]$.
\STATE Calculate $h_j$ and $\omega_j$ by Eq. (53)-(54) and select $j^*$ by $\min \{ \omega_j \}_{1 \leq j \leq T_{\max}}$.
\STATE Save $w_{j^*}$ in $W$ and $b_{j^*}$ in $B$.
\STATE Calculate $h_{j^*}^{\dagger}$ and store as $H_1^{\dagger}$.
\STATE $e_1 = H_L H_L^{\dagger} Y - Y$.
\STATE $L = 1$.

\WHILE{$L \leq L_{\max}$ \AND $\| e_{L-1} \|_F > \epsilon$}
    \STATE Set $r^* = r^{(1 + 1/L)^\gamma}$ and load $H_{L-1}^{\dagger}$.
    \FOR{$\lambda \in \Upsilon$}
        \STATE Randomly assign input weights $w_j$ and $b_j$, $1 \leq j \leq T_{\max}$ from $[-\lambda, \lambda]^d$ and $[-\lambda, \lambda]$.
        \STATE Calculate $h_j$, $p_j$, and $\xi_j$ based on Eq. (53) and Eq. (55)-(56).
        \IF{$\min \{\xi_j\}_{1 \leq j \leq T_{\max}} \leq 0$}
            \STATE Save $w_{j^*}$ in $W$ and $b_{j^*}$ in $B$ where $j^*$ is selected by $\min \{\xi_j\}_{1 \leq j \leq T_{\max}}$.
            \STATE Set $H_L = [H_{L-1}, h_{j^*}]$ and update $H_L^{\dagger}$ by Eq. (60).
            \STATE $e_L = H_L H_L^{\dagger} T - T$.
            \STATE $L = L + 1$.
            \STATE \textbf{break}
        \ENDIF
    \ENDFOR
\ENDWHILE
\STATE \textbf{return} $W,B$
\end{algorithmic}
\end{algorithm}

\noindent Given a basis function $g = \sigma(w^T x + b)$, $x \in \mathbb{R}^d$ with the input weight $w^T = [w_1, \dots, w_d]$ and bias $b$,
\begin{equation}
    \label{eq53}
    \begin{array}{cc}
         { h = g(X) = }  \\
         {\begin{bmatrix} \sigma(w^T x_1 + b_j), \dots, \sigma(w^T x_N + b_j) \end{bmatrix}^T, h \in \mathbb{R}^N. } 
    \end{array}
\end{equation}
Denote the $q$-th column of $Y$ as $Y_q$ where $Y_q = \begin{bmatrix} y_{1,q}, y_{2,q}, \dots, y_{N,q} \end{bmatrix}^T$.

\noindent For the initialization of RMPI-SCN, $h_j$ is evaluated through $\omega_j$ where 

\begin{equation}
    \label{eq54}
    \omega_j = \sum_{q=1}^m \left\| \left( \frac{Y_q^T \cdot h_j}{h_j^T \cdot h_j} \right) \cdot h_j - Y_q \right\|.
\end{equation}
Let $H_{L-1} = [h_1, h_2, \dots, h_{L-1}]$, $H_{L-1} \in \mathbb{R}^{N \times (L-1)}$ and $H_{L-1}$ is the output of the current hidden layer on $X$. $H_{L-1}^{\dagger}$ is the Moore-Penrose inverse of $H_{L-1}$. Denote the current training residual as $e_{L-1}$ where $e_{L-1} = [e_{L-1,1}, \dots, e_{L-1,m}]$, $e_{L-1} \in \mathbb{R}^{N \times m}$ and $e_{L-1,q} = H_{L-1} H_{L-1}^{\dagger} Y_q - Y_q$, $e_{L-1,q} \in \mathbb{R}^N$, $1 \leq q \leq m$.
\begin{equation}
    \label{eq55}
    p_j = h_j - H_{L-1} \cdot H_{L-1}^{\dagger} \cdot h_j.
\end{equation}
For the calculation of $\xi_j$,
\begin{equation}
    \label{eq56}
    \xi_j = \sum_{q=1}^m \xi_{j,q} - r^* \cdot \sum_{q=1}^m e_{L-1,q}^T \cdot e_{L-1,q}, 
\end{equation}
and
\begin{equation}
    \label{eq57}
    \xi_{j,q} = \frac{|p_j^T \cdot Y_q|^2}{p_j^T \cdot p_j} + 2 \cdot \frac{(p_j^T \cdot Y_q)(e_{L-1,q}^T \cdot p_j)}{p_j^T \cdot p_j} + e_{L-1,q}^T \cdot e_{L-1,q}.
\end{equation}
The recursive calculation of $H_L^{\dagger}$ is:
\begin{equation}
    \label{eq58}
    d_L = H_{L-1}^{\dagger} \cdot h_L,
\end{equation}
\begin{equation}
    \label{eq59}
    b_L = \begin{cases} 
      (1 + d_L^T d_L)^{-1} \cdot d_L^T \cdot H_{L-1}^{\dagger}, & p_L = 0, \\
      p_L^{\dagger}, & p_L \neq 0,
   \end{cases}
\end{equation}

\begin{equation}
    \label{eq60}
    H_L^{\dagger} = \begin{bmatrix} H_{L-1}^{\dagger} - d_L \cdot b_L \\ b_L \end{bmatrix}.
\end{equation}

\section{	Simulation results}
\noindent This section reports the performance of Recursive Moore-Penrose Inverse SCN (RMPI-SCN) across ten regression tasks, including two function approximation examples [12, 28] and eight benchmark datasets from UCI and KEEL. The benchmark datasets were downloaded from the KEEL and UCI repositories. Each dataset is divided into training, validation, and test sets in a ratio of 6:2:2, followed by normalization to ensure consistent scales across features. The performance of RMPI-SCN is compared against SCN-III, RVFL, and a Multi-Layer Perceptron (MLP) trained using Backpropagation (BP). Hyperparameters for each model were selected through cross-validation. In particular, the number of nodes for the MLP was chosen from the set {5, 10, 15, 20, 25, 30, 50} using cross-validation techniques. An early stopping strategy was implemented to evaluate model performance before overfitting on the validation sets. Each model's performance was assessed through 100 independent experiments, employing the logistic sigmoid function as the activation function. The functions for the approximations, DB1 and DB2, are described as follows:

\noindent DB1: DB1 consists of 1500 samples from a regularly spaced grid on [0,1], drawn from
\begin{equation}
\label{eq61}
f(x) = 0.2 e^{-(10x - 4)^2} + 0.5 e^{-(80x - 40)^2} + 0.3 e^{-(80x - 20)^2}. 
\end{equation}
DB2: DB2 is sampled from a classic nonlinear model in system identification given by:
\begin{equation}
\label{eq62}
y(k+1) = f[y(k), y(k-1), y(k-2), u(k), u(k-1)],
\end{equation}
where
\begin{equation}
\label{eq63}
f(x_1, x_2, x_3, x_4, x_5) = \frac{x_1 x_2 x_3 x_5 (x_4 - 1)}{1 + x_2^2 + x_3^2}. 
\end{equation}
The training samples and validation samples of DB2 are generated through 2400 iterations of random signals $u$ uniformly distributed over $[-1,1]$. The 600 test samples for DB2 are obtained through the function:
\begin{equation}
    \label{eq64}
u(k) = 
\begin{cases}

\sin\left(\frac{2 \pi k}{250}\right), & 1 \leq k \leq 250, \\
0.8 \sin\left(\frac{2 \pi k}{250}\right) + 0.2 \sin\left(\frac{2 \pi k}{25}\right), & k > 250.
\end{cases} 
\end{equation}
The datasets DB3-DB10 are benchmark datasets: CCPP, Superconduct, Delta Ail, Stock, Concrete, and CCPP, downloaded from KEEL and UCI.
\begin{table}[h]
\centering
\resizebox{\columnwidth}{!}{ % Resize table to fit column width
\begin{tabular}{cccccccc}
\hline
\textbf{Names} & \textbf{Function} & \textbf{Sample Size} & \textbf{Training} & \textbf{Validation} & \textbf{Test} & \textbf{Input} & \textbf{Output} \\
 &  &  & \textbf{Samples} & \textbf{Samples} & \textbf{Samples} & \textbf{Variables} & \textbf{Variables} \\
\hline
DB1 & Eq.(\ref{eq61}) & 1500 & 900 & 300 & 300 & 1 & 1 \\
\hline
DB2 & Eq.(\ref{eq63}) & 3000 & 1800 & 600 & 600 & 5 & 1 \\
\hline
DB3 & CCPP & 9568 & 5740 & 1913 & 1915 & 4 & 1 \\
\hline
DB4 & Superconduct & 21263 & 12757 & 4252 & 4254 & 81 & 1 \\
\hline
DB5 & Delta ail & 7129 & 4277 & 1425 & 1427 & 5 & 1 \\
\hline
DB6 & Friedman & 1200 & 720 & 240 & 240 & 5 & 1 \\
\hline
DB7 & House & 22784 & 13670 & 4456 & 4458 & 16 & 1 \\
\hline
DB8 & Concrete & 1030 & 618 & 206 & 206 & 6 & 1 \\
\hline
DB9 & Stock & 950 & 570 & 190 & 190 & 9 & 1 \\
\hline
DB10 & Wizmir & 1461 & 876 & 292 & 293 & 9 & 1 \\
\hline
\end{tabular}
}
\caption{The descriptive statistics of DB1-DB10}
\end{table}

\noindent For the comparison of effectiveness, we report the Root Mean Square Error (RMSE) and correlation coefficient (R). The efficiency is measured by the time of model construction. 

\noindent The RMSE is calculated using the formula:
\begin{equation}
\label{eq65}
\text{RMSE} = \sqrt{\frac{1}{n} \sum_{i=1}^n \| \hat{y}_i - y_i \|^2}, 
\end{equation}
where $\hat{y}_i$ represents the predicted values and $y_i$ denotes the actual values.

\noindent The correlation coefficient ($R$) is defined as:
\begin{equation}
\label{eq66}
R = \frac{\text{Cov}(y, \hat{y})}{\sqrt{\text{Var}(y)} \sqrt{\text{Var}(\hat{y})}}, 
\end{equation}
where $\text{Cov}$ denotes the covariance and $\text{Var}$ represents the variance of the actual and predicted values.

\begin{table*}[ht]
\centering
\caption{Simulation results of RMPI-SCN, SCN-III, RVFL and MLP-BP over DB1-DB10}
\label{tab3}
\setlength{\tabcolsep}{4pt}
\begin{tabular}{@{}cccccccc@{}}
\toprule
Data & Algorithm & Train RMSE & Train R & Test RMSE & Test R & Training Time(s) & Nodes \\ \midrule
\multirow{4}{*}{DB1} & RMPI-SCN & $\mathbf{0.0014\pm0.0009}$ & $\mathbf{0.9998\pm0.0003}$ & $\mathbf{0.0016\pm0.0010}$ & $\mathbf{0.9998\pm0.0003}$ & $0.5093\pm0.0719$ & 45.1 \\
 & SCN-III & $0.0029\pm0.0023$ & $0.9992\pm0.0013$ & $0.0032\pm0.0025$ & $0.9992\pm0.0012$ & $0.5521\pm0.0642$ & 57.9 \\
 & RVFL & $0.0441\pm0.0002$ & $0.8743\pm0.0029$ & $0.0515\pm0.0004$ & $0.8607\pm0.0020$ & $\mathbf{0.0034\pm0.0014}$ & 100 \\
 & MLP-BP & $0.0026\pm0.0064$ & $0.9972\pm0.0125$ & $0.0030\pm0.0074$ & $0.9972\pm0.0128$ & $0.8521\pm0.4259$ & 20 \\
\midrule
\multirow{4}{*}{DB2} & RMPI-SCN & $\mathbf{0.0059 \pm 0.0004}$ & $\mathbf{0.9999 \pm 0.0000}$ & $\mathbf{0.0344 \pm 0.0053}$ & $\mathbf{0.9980 \pm 0.0005}$ & $0.4878 \pm 0.0132$ & 119.6 \\
 & SCN-III & $0.0083 \pm 0.0005$ & $0.9998 \pm 0.0000$ & $0.0537 \pm 0.0011$ & $0.9951 \pm 0.0019$ & $0.4215 \pm 0.0459$ & 119.8 \\
 & RVFL & $0.0203 \pm 0.0018$ & $0.9989 \pm 0.0002$ & $0.0628 \pm 0.0196$ & $0.9928 \pm 0.0044$ & $\mathbf{0.0052 \pm 0.0004}$ & 120 \\
 & MLP-BP & $0.0079 \pm 0.0027$ & $0.9998 \pm 0.0001$ & $0.0394 \pm 0.0276$ & $0.9960 \pm 0.0077$ & $0.8979 \pm 0.3536$ & 15 \\ 
 \midrule
 \multirow{4}{*}{DB3} & RMPI-SCN & $\mathbf{3.7412\pm0.0272}$ & $\mathbf{0.9841\pm0.0004}$ & $\mathbf{3.9136\pm0.0243}$ & $\mathbf{0.9771\pm0.0003}$ & $0.6268\pm0.0224$ & 65.4 \\
 & SCN-III & $3.9124\pm0.0183$ & $0.9730\pm0.0003$ & $3.9635\pm0.0171$ & $0.9742\pm0.0002$ & $0.8955\pm0.0269$ & 69.3 \\
 & RVFL & $4.0662\pm0.0151$ & $0.9708\pm0.0002$ & $4.0980\pm0.0205$ & $0.9724\pm0.0003$ & $\mathbf{0.0123\pm0.0008}$ & 70 \\
 & MLP-BP & $3.9225\pm0.0482$ & $0.9729\pm0.007$ & $3.9696\pm0.0466$ & $0.9741\pm0.0006$ & $0.7117\pm0.1259$ & 30 \\
 \midrule
\multirow{4}{*}{DB4} & RMPI-SCN & $\mathbf{11.4750\pm0.4574}$ & $\mathbf{0.9425\pm0.0047}$ & $\mathbf{12.4767\pm0.2500}$ & $\mathbf{0.9306\pm0.0029}$ & $43.031\pm21.826$ & 268.3 \\
 & SCN-III & $12.2079\pm0.0516$ & $0.9348\pm0.0006$ & $13.1137\pm0.0925$ & $0.9230\pm0.0011$ & $24.3728\pm0.7779$ & 398.3 \\
 & RVFL & $14.7074\pm0.0838$ & $0.9039\pm0.0012$ & $15.1686\pm0.1007$ & $0.8954\pm0.0015$ & $\mathbf{0.4347\pm0.0301}$ & 400 \\
 & MLP-BP & $12.1200\pm0.3880$ & $0.9358\pm0.0042$ & $13.1899\pm0.2068$ & $0.9224\pm0.0025$ & $52.942\pm11.0015$ & 50 \\
 \midrule
\multirow{4}{*}{DB5} & RMPI-SCN & $1.6605e^{-4}\pm1.4833e^{-6}$ & $0.8433\pm0.0031$ & $\mathbf{1.5809e^{-4}\pm8.5239e^{-7}}$ & $\mathbf{0.8486\pm0.0018}$ & $0.5302\pm0.1247$ & 24.9 \\
 & SCN-III & $
 \mathbf{1.6581e^{-4}\pm1.9459e^{-6}}$ & $\mathbf{0.8468\pm0.0041}$ & $1.6260e^{-4}\pm1.6255e^{-6}$ & $0.8392\pm0.0035$ & $0.5054\pm0.0204$ & 39.1 \\
 & RVFL & $1.6953e^{-4}\pm8.1211e^{-7}$ & $0.8360\pm0.0017$ & $1.6059e^{-4}\pm1.2586e^{-6}$ & $0.8433\pm0.0026$ & $\mathbf{0.0063\pm0.0012}$ & 50 \\
 & MLP-BP & $1.6687e^{-4}\pm2.1296e^{-6}$ & $0.8419\pm0.0043$ & $1.6001e^{-4}\pm2.8199e^{-6}$ & $0.8451\pm0.0054$ & $0.5670\pm0.0596$ & 10 \\ 
 \midrule
 \multirow{4}{*}{DB6} & RMPI-SCN & $\mathbf{0.9707\pm0.0153}$ & $\mathbf{0.9827\pm0.0011}$ & $1.3385\pm0.0531$ & $\mathbf{0.9647\pm0.0028}$ & $0.1600\pm0.0230$ & 57.3 \\
 & SCN-III & $1.0670\pm0.0482$ & $0.9790\pm0.0020$ & $1.4321\pm0.0553$ & $0.9595\pm0.0032$ & $0.1295\pm0.0014$ & 62.5 \\
 & RVFL & $1.6097\pm0.1193$ & $0.9513\pm0.0074$ & $1.9211\pm0.1586$ & $0.9252\pm0.0129$ & $\mathbf{0.0015\pm0.0031}$ & 70 \\
 & MLP-BP & $1.2004\pm0.2749$ & $0.9720\pm0.0153$ & $\mathbf{1.3043\pm0.2883}$ & $0.9646\pm0.0189$ & $0.5288\pm0.1162$ & 5 \\
 \midrule
\multirow{4}{*}{DB7} & RMPI-SCN & $\mathbf{33359\pm224.00}$ & $\mathbf{0.7751\pm0.0035}$ & $\mathbf{37580\pm339.47}$ & $0.7046\pm0.0059$ & $5.4222\pm0.8891$ & 193.5 \\
 & SCN-III & $34011\pm182.58$ & $0.7648\pm0.0029$ & $37892\pm291.18$ & $0.6979\pm0.0052$ & $5.3493\pm0.2413$ & 195.02 \\
 & RVFL & $39091\pm68.92$ & $0.6721\pm0.0056$ & $40586\pm295.58$ & $0.6379\pm0.0066$ & $\mathbf{0.1171\pm0.0092}$ & 200 \\
 & MLP-BP & $34570\pm1078.44$ & $0.7559\pm0.0181$ & $37604\pm802.91$ & $\mathbf{0.7063\pm0.0142}$ & $6.6683\pm0.2345$ & 30 \\
 \midrule
\multirow{4}{*}{DB8} & RMPI-SCN & $\mathbf{4.4220\pm0.2642}$ & $\mathbf{0.9639\pm0.0045}$ & $\mathbf{6.1905\pm0.1836}$ & $\mathbf{0.9300\pm0.0042}$ & $0.3268\pm0.0253$ & 87.61 \\
 & SCN-III & $4.8125\pm0.3487$ & $0.9570\pm0.0066$ & $6.4926\pm0.1559$ & $0.9231\pm0.0037$ & $0.2593\pm0.0106$ & 84.88 \\
 & RVFL & $6.6387\pm0.2381$ & $0.9168\pm0.0062$ & $8.0629\pm0.4721$ & $0.8785\pm0.01398$ & $\mathbf{0.0017\pm0.0002}$ & 100 \\
 & MLP-BP & $4.9908\pm0.5138$ & $0.9542\pm0.0099$ & $6.6900\pm0.6973$ & $0.9193\pm0.0153$ & $0.5340\pm0.0607$ & 25 \\
 \midrule
 \multirow{4}{*}{DB9} & RMPI-SCN & $\mathbf{0.4532\pm0.0446}$ & $\mathbf{0.9975\pm0.0005}$ & $\mathbf{0.8009\pm0.0472}$ & $\mathbf{0.9927\pm0.0009}$ & $0.3806\pm0.0254$ & 101.9 \\
 & SCN-III & $0.5575\pm0.0471$ & $0.9962\pm0.0006$ & $0.8103\pm0.0426$ & $0.9925\pm0.0008$ & $0.2212\pm0.0015$ & 101.7 \\
 & RVFL & $0.7677\pm0.0299$ & $0.9928\pm0.0006$ & $0.9202\pm0.0577$ & $0.9903\pm0.0013$ & $\mathbf{0.0017\pm0.0002}$ & 100 \\
 & MLP-BP & $0.6966\pm0.0628$ & $0.9942\pm0.0010$ & $0.8807\pm0.0667$ & $0.9912\pm0.0014$ & $0.4757\pm0.0556$ & 30 \\
 \midrule
\multirow{4}{*}{DB10} & RMPI-SCN & $\mathbf{0.8915\pm0.0232}$ & $\mathbf{0.9981\pm0.0001}$ & $1.4124\pm0.0510$ & $0.9949\pm0.0004$ & $0.3336\pm0.1025$ & 88.40 \\
 & SCN-III & $0.9901\pm0.0764$ & $0.9977\pm0.0004$ & $1.4618\pm0.0594$ & $0.9946\pm0.0004$ & $0.1979\pm0.0017$ & 79.8 \\
 & RVFL & $0.9988\pm0.0152$ & $0.9976\pm0.0001$ & $\mathbf{1.3750\pm0.0392}$ & $\mathbf{0.9952\pm0.0003}$ & $\mathbf{0.0022\pm0.0003}$ & 100 \\
 & MLP-BP & $1.0004\pm0.0572$ & $0.9975\pm0.0003$ & $1.6140\pm0.4030$ & $0.9930\pm0.0042$ & $0.4296\pm0.0140$ & 20 \\ 
% ... 其他数据行 ...
\bottomrule
\end{tabular}
\end{table*}

\noindent Table \ref{tab3} summarize the training and testing performance of RMPI-SCN, SCN-III, RVFL and MLP trained by BP across the datasets DB1-DB10. Notably, RMPI-SCN demonstrated the lowest average training RMSE and the highest correlation coefficients on nine out of ten datasets, with DB5 being the only exception where the difference in training RMSE between RMPI-SCN and SCN-III was negligible. RMPI-SCN particularly excelled in function approximation tasks, achieving average training RMSEs of 0.0014 for DB1 and 0.0059 for DB2, compared to the second-best algorithm, MLP-BP, which recorded RMSEs of 0.0026 and 0.0079, respectively. Similar performance trends are observed in the benchmark datasets, including DB3, DB4, DB6, and DB8-10. 

\noindent Furthermore, both RMPI-SCN and SCN-III exhibited more stable learning performances compared to MLP-BP, as evidenced by the smaller standard deviations in training RMSE and correlation coefficients. For example, the standard deviations of training RMSE for RMPI-SCN and SCN-III on DB7 were 224.00 and 182.58, respectively, while MLP-BP had a substantially higher standard deviation of 1078.44. This consistency in performance across various datasets underscores the strong learning capabilities of RMPI-SCN. Overall, the statistics suggest that RMPI-SCN offers significant advantages over the classic SCN-III and MLP-BP, particularly in function approximation tasks, thus providing a robust framework for improved modeling performance.

\noindent Regarding generalization performance, RMPI-SCN achieved the lowest testing RMSE on datasets DB1-5 and DB7-9. Notably, the testing performances on DB1, DB2, and DB8 showed significant differences, with average RMSE values of 0.0016, 0.0344, and 6.1905 for RMPI-SCN, respectively. In comparison, the average RMSE of the second-best algorithms on these datasets were 0.0030 (MLP-BP), 0.0394 (MLP-BP), and 6.4926 (SCN-III). These findings collectively indicate that RMPI-SCN maintains stable generalization performance on test sets.

\noindent Tables \ref{tab3} also clearly illustrate that the three randomized learning algorithms—RMPI-SCN, SCN-III, and RVFL—exhibit significant superiority over MLP-BP in terms of efficiency across nearly all datasets. The only exception is DB5, where the differences in training times among RMPI-SCN, SCN-III, and MLP-BP are negligible. Among the three randomized algorithms, RVFL consistently demonstrated lower training times compared to RMPI-SCN and SCN-III across all datasets. However, RVFL's learning and generalization performances were only satisfactory on DB3, DB5, and DB10. In the function approximation tasks, DB1 and DB2, RVFL's training and testing RMSEs were notably higher than those of the other algorithms, indicating its limited learning capability.

\noindent When comparing RMPI-SCN to the original SCN-III, there is a moderate increase in training time overall. The exception is DB4, where the average training time for RMPI-SCN is 43.0383 seconds, whereas SCN-III's training time is only 24.3728 seconds. Although RMPI-SCN eliminates the inner searching loop for r and utilizes a recursive method to calculate the Moore-Penrose inverse of the output matrix from the hidden layer, its increased sensitivity to expanding the support of uniform distributions inevitably results in higher computational costs.

\begin{figure}[htbp]
    \centering
    \includegraphics[width=1\linewidth]{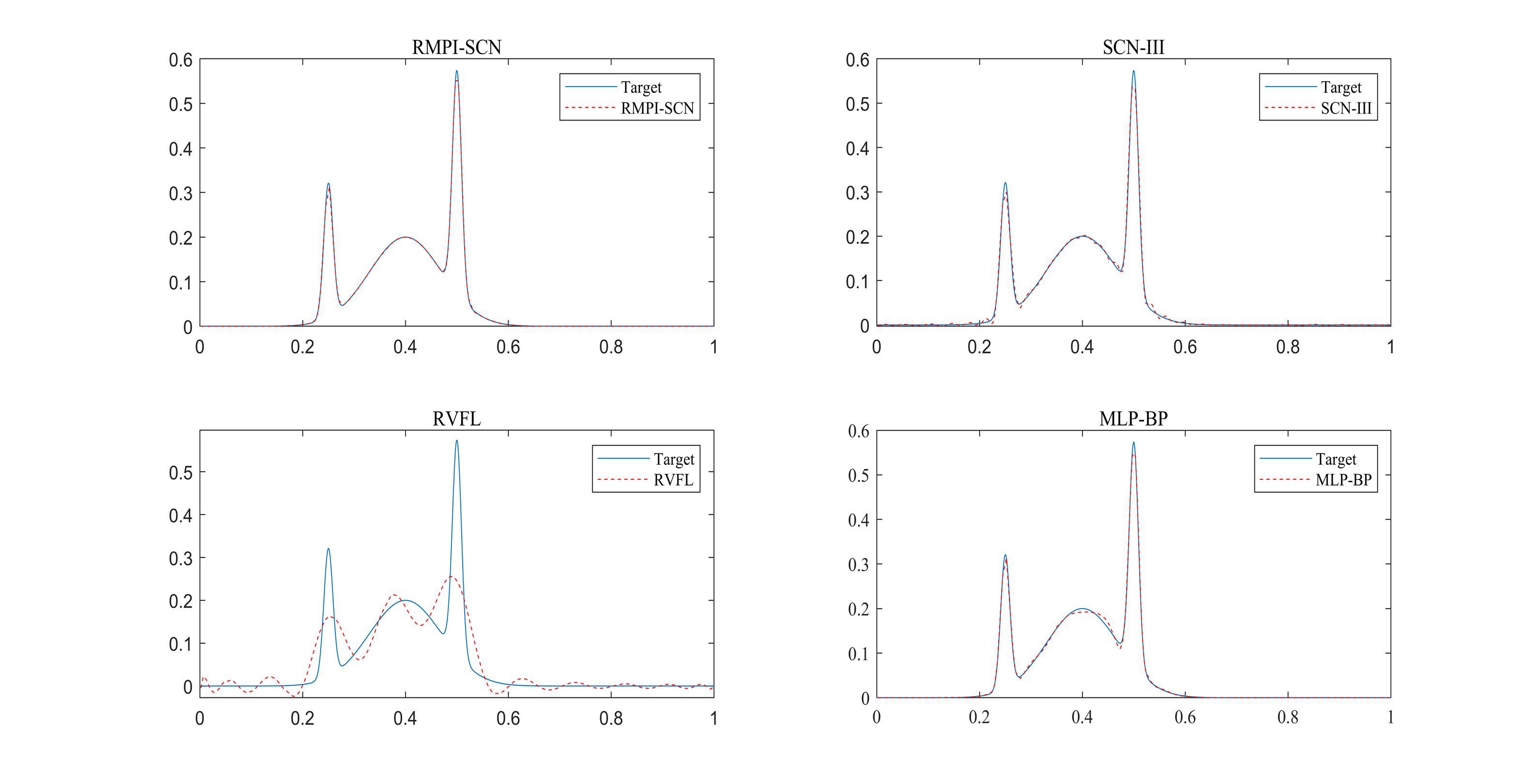}
    \caption{Approximation performance of RMPI-SCN, SCN-III, RVFL and MLP-BP on DB1}
    \label{fig3}
\end{figure}

\noindent To further illustrate the characteristics of RMPI-SCN and SCN-III, we present a detailed discussion of the simulations conducted on DB1. Figure \ref{fig3} shows the approximations of RMPI-SCN, SCN-III, MLP-BP and RVFL on DB1.

\noindent The results indicate satisfactory approximations for all methods except RVFL. Notably, RMPI-SCN significantly outperforms SCN-III and MLP-BP, achieving a training RMSE of 2.8254e-04 and a testing RMSE of 3.4588e-04. In contrast, SCN-III's best performance in the 100 independent experiments resulted in a training RMSE of 0.0016 and a testing RMSE of 0.0020. In comparison to the other models, RVFL struggles to capture these abrupt changes in the target function accurately, which indicates limitations in its ability to represent complex, high-frequency components of the function. The experiments on DB1 highlights one of the key limitations of RVFL: its inability to adaptively capture the derivative of the target function. This limitation stems from the fact that RVFL’s basis functions are randomly generated from a fixed uniform distribution, lacking any adaptive mechanism to align these functions with the finer details of the target function, including its derivatives. 

\begin{figure}[htbp]
    \centering
    \includegraphics[width=1\linewidth]{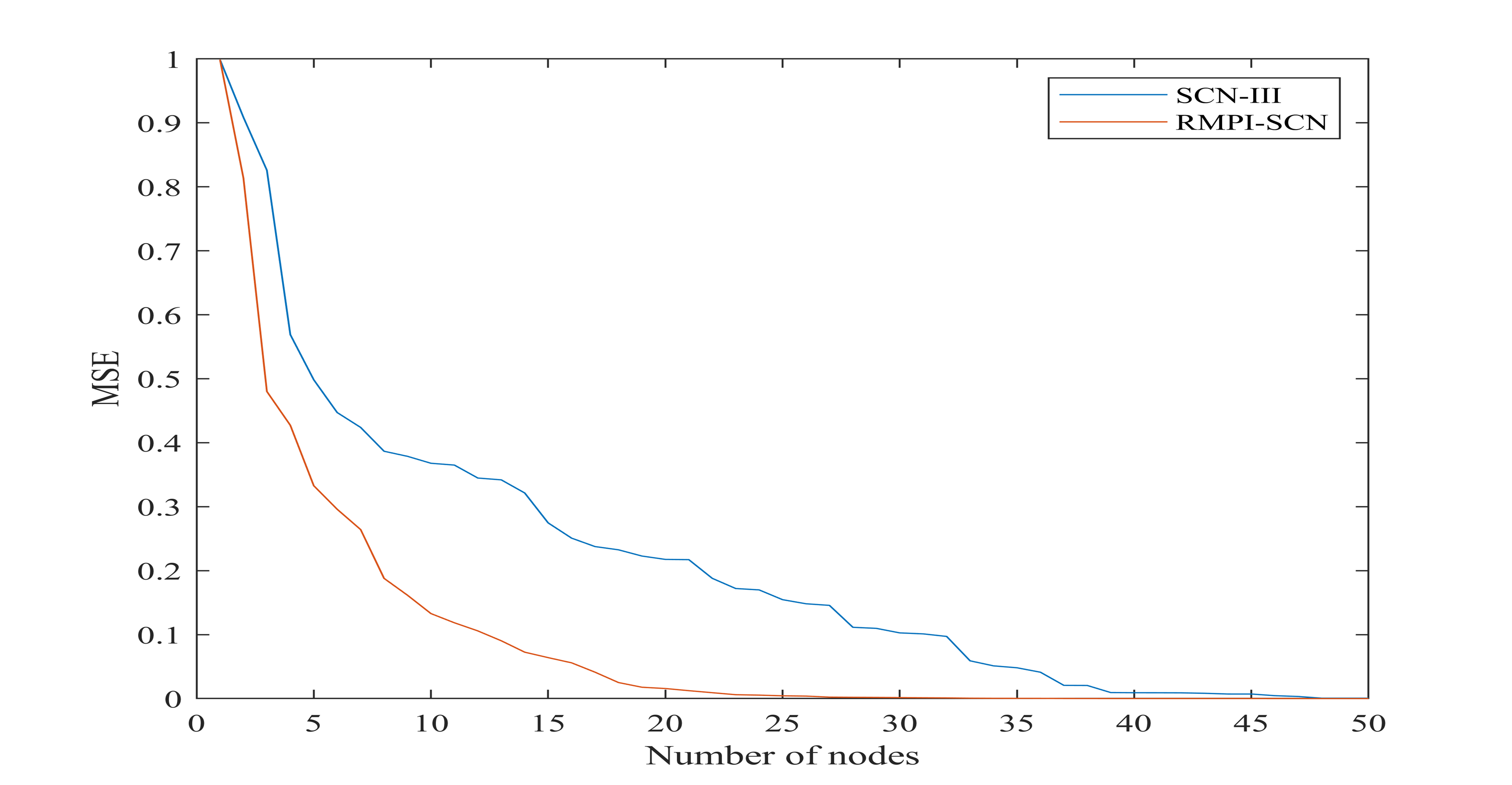}
    \caption{Training errors of RMPI-SCN and SCN-III on DB1}
    \label{fig4}
\end{figure}

\begin{figure}[htbp]
    \centering
    \includegraphics[width=1\linewidth]{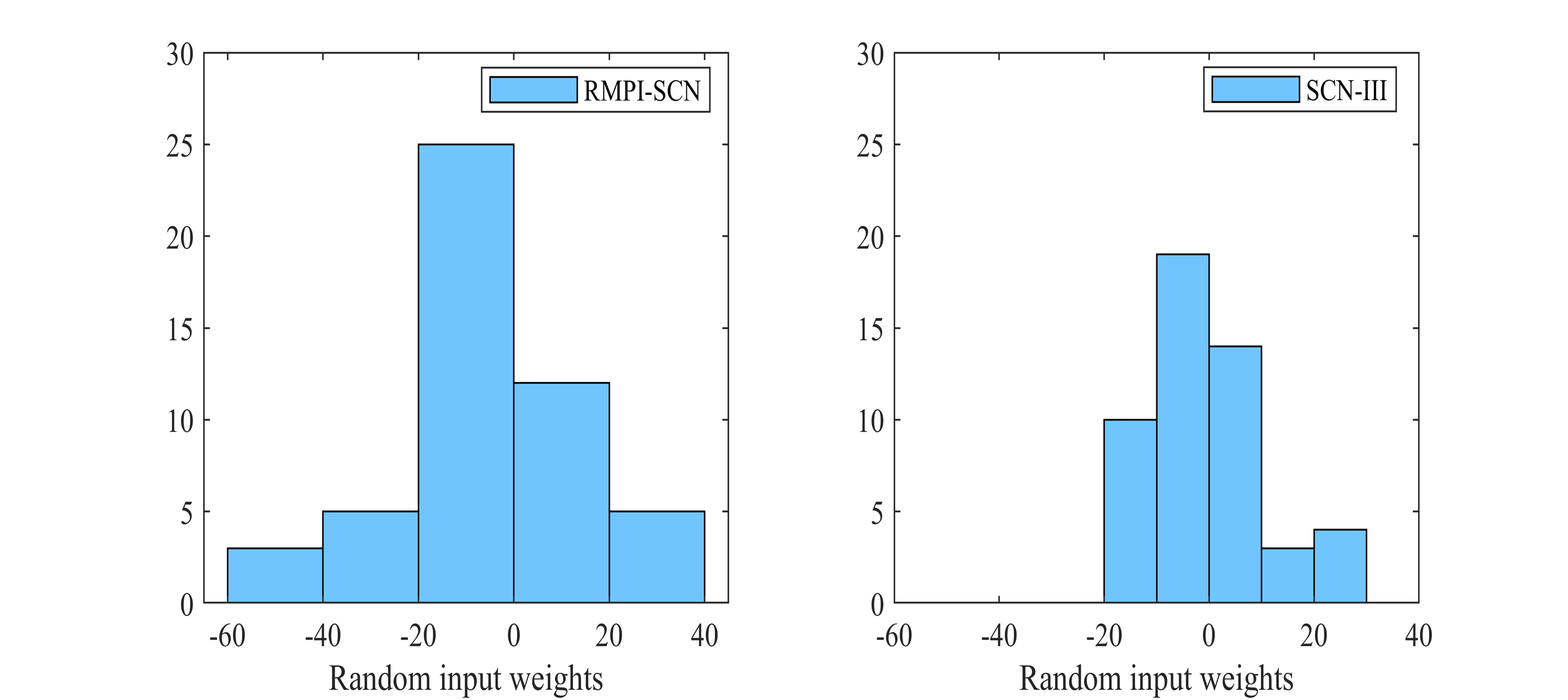}
    \caption{The histogram of random input weights of RMPI-SCN and SCN-III on DB1}
    \label{fig5}
\end{figure}

\noindent Figure \ref{fig4} illustrates the training error reduction curves for RMPI-SCN and SCN-III. The RMPI-SCN converges significantly faster than SCN-III, with consistently lower training residual errors throughout the simulation. This phenomenon is not an experiment coincidence but a mathematical consequence. Since the RMPI-SCN evaluates the random basis function by $\|Y-H_L H_L^{\dagger} Y\|$ as defined by Eq. (\ref{eq52}) in Corollary 3. This approach ensures that RMPI-SCN selects random basis functions that most effectively lower the residual error at each iteration. We further analyze the distribution of random input weights in Figure \ref{fig5}.

\noindent The central concept of SCN is to adaptively expand the support of the uniform distribution based on data and current training residuals. Since the inequality constraints in Eq. (50)-(51) in Corollary 3 of RMPI-SCN provide a necessary and sufficient condition for $\| Y - H_L H_L^{\dagger} Y \| \leq \sqrt{r} \| e_{L-1} \|$
This improvement allows for more accurate assessments of whether the current uniform distribution can generate suitable random basis functions compared to the original constraint in Eq. (\ref{eq4}) in SCN-III. As a result, the learning rate $r$ can be set to smaller values, providing higher sensitivity for expanding the uniform distribution. Consequently, RMPI-SCN's learning rate $r$ can be set to smaller values, enhancing its sensitivity for expanding the uniform distribution.

\noindent We also examine the limitations of RVFL and emphasize the necessity of adaptively adjusting the distribution for random input weights in randomized learning. Figure \ref{fig7} depicts the target function of DB1, alongside its first and second derivatives. The second derivative ranges from $[-6000, 2000]$ around $x = 0.5$, which is a key factor in the failure of RVFL. For a basis function $g$ based on a sigmoid activation function, with input weights $w$ and bias $b$, the second derivative can be expressed as $w^2 \frac{1 - e^{wx + b}}{1 + e^{wx + b}}. $ Given that $\left|\frac{1 - e^{wx + b}}{1 + e^{wx + b}}\right|$ is bounded by one, if the input weight $w$ is randomly generated from a uniform distribution within $[-1, 1]$, it becomes clear that approximating the second derivative of the target function requires at least 6000 such basis functions.

\begin{figure}[htbp]
    \centering
    \includegraphics[width=1\linewidth]{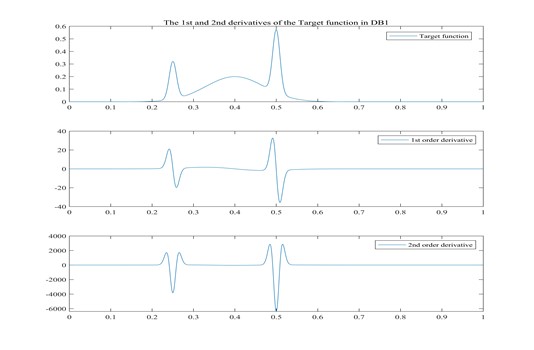}
    \caption{The target function of DB1 and its 1st and 2nd order derivatives in (0,1)}
    \label{fig6}
\end{figure}

\noindent An inaccurate approximation of the second-order derivative consequently leads to significant discrepancies in the first-order derivative, resulting in approximation failure. Additionally, assigning large output weights to some basis functions is not feasible, as it dramatically increases the overall output of the basis function. To mitigate these issues, we adjusted the support of the uniform distribution for RVFL to [-60, 60] and present the updated approximation results in Figure \ref{fig7}. This case study on DB1 underscores the importance of adaptively adjusting the distribution for generating random inputs and biases, aligning with the core motivation of SCNs. 

\begin{figure}[htbp]
    \centering
    \includegraphics[width=1\linewidth]{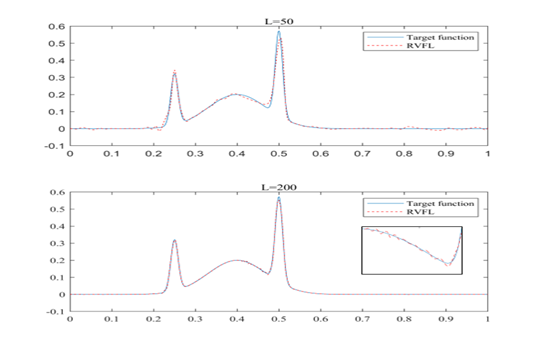}
    \caption{The approximation of RVFL where the support of the uniform distribution is [-60,60]}
    \label{fig7}
\end{figure}

\begin{figure}[htbp]
    \centering
    \includegraphics[width=1\linewidth]{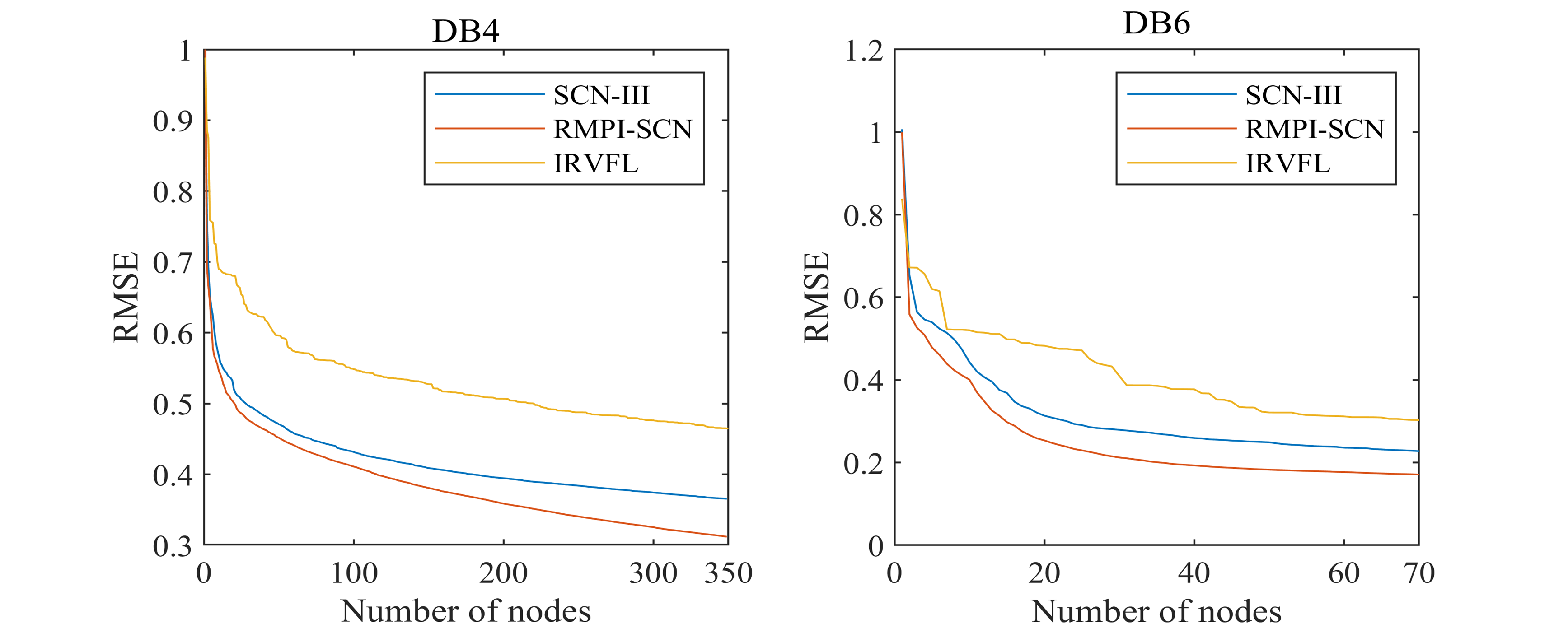}
    \caption{Training error reduction curve of IRVFL, RMPI-SCN and SCN-III on DB4 and DB6}
    \label{fig8}
\end{figure}

\begin{figure}[htbp]
    \centering
    \includegraphics[width=1\linewidth]{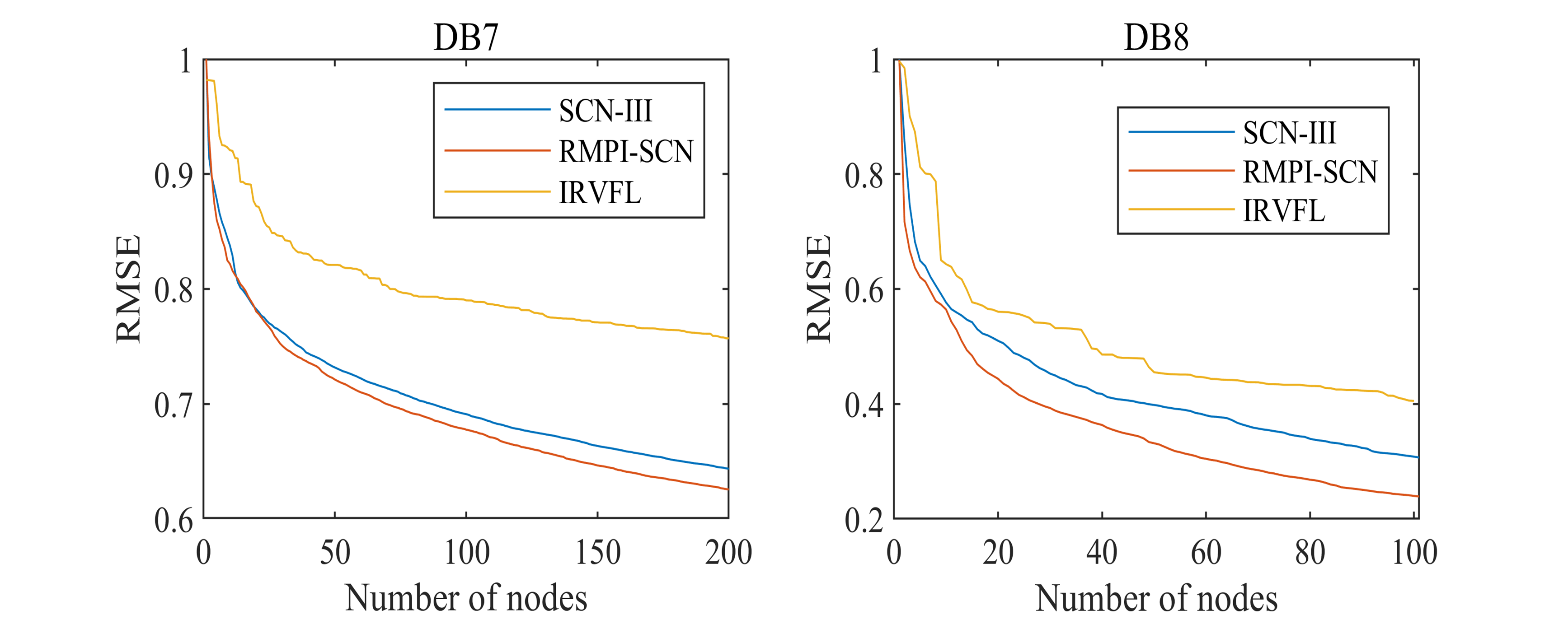}
    \caption{Training error reduction curve of IRVFL, RMPI-SCN and SCN-III on DB7 and DB8}
    \label{fig9}
\end{figure}

\noindent Figures \ref{fig8} and \ref{fig9} illustrate the training error reduction curves for DB4, DB6, and DB7-8, further supporting the efficacy of RMPI-SCN in various contexts. To facilitate comprehensive comparison, these figures also include the error reduction curve for Incremental Random Vector Functional Links (IRVFL). As shown, RMPI-SCN’s training residual error converges faster and more smoothly than both SCN-III and IRVFL, consistently achieving lower RMSE with fewer nodes. This superior performance underscores the efficacy of RMPI-SCN in achieving rapid error reduction, a crucial aspect in contexts requiring efficient learning and precise function approximation. Notably, the supervised nature of both SCN-III and RMPI-SCN allows these models to dynamically evaluate and select basis functions based on their contributions to reducing residual error, setting them apart from the unsupervised IRVFL model. In contrast, IRVFL lacks this level of supervision, relying solely on randomized basis function selection without iterative error-based refinement. Consequently, IRVFL’s error reduction curve demonstrates a slower and less consistent convergence pattern, as shown in Figures \ref{fig8} and \ref{fig9}.

\noindent Moreover, RMPI-SCN’s training residual error converges faster and more smoothly than both SCN-III and IRVFL, indicating that RMPI-SCN consistently achieves lower RMSE with fewer nodes, underscoring its effectiveness in achieving rapid error reduction. For instance, in DB4, SCN-III shows minimal reductions in training errors as the number of hidden nodes increases from 250 to 300, as indicated by the near-flat curve toward the end of the training process. In contrast, RMPI-SCN continues to demonstrate significant decreases in training error with the expansion of network size. This pattern is also evident in the comparisons with IRVFL, where RMPI-SCN maintains a lower and more consistently decreasing RMSE across all node counts, showcasing its robust convergence capability relative to IRVFL.

\noindent Furthermore, RMPI-SCN achieves these results with fewer nodes on average. For example, on DB4, RMPI-SCN requires an average of 268 nodes, whereas SCN-III averages 398 nodes. Additionally, RMPI-SCN attains an average training RMSE and testing RMSE of 11.475 and 12.4767, respectively, compared to 12.2079 and 13.1137 for SCN-III. These differences underscore RMPI-SCN’s efficiency and accuracy in reducing residual errors. DB4 serves as a concrete example highlighting the drawbacks of the current supervisory mechanism of SCN-III. Since the value of ξ given in Eq. (\ref{eq10}) assumes that the output weights for previously added basis functions remain unchanged, it creates a strict lower bound for the capabilities of these basis functions to reduce training errors. Consequently, r must be set very close to one to ensure that there are random basis functions. This limitation hampers SCN-III's ability to adjust the uniform distribution for random input weights and biases based on the reduction of training residuals, ultimately constraining its learning capability.


\noindent The simulations conducted on DB1 and DB4 provide strong evidence that accurate evaluations of the current distribution’s capacity to generate suitable random basis functions are crucial for the effectiveness of randomized learning models. The findings affirm that an adaptive approach to adjusting the support of the uniform distribution leads to improved model performance, particularly in tasks requiring precise function approximation.

\noindent In summary, the presented simulation results illustrate the effectiveness and efficiency of the proposed RMPI-SCN algorithm, showcasing its superior performance in training and generalization across various datasets. The comparison with SCN-III, RVFL, and MLP-BP reinforces the advantages of RMPI-SCN, particularly in terms of stability, learning capability, and computational efficiency. These insights provide a comprehensive understanding of the strengths of RMPI-SCN and highlight the importance of adaptively managing random input distributions in achieving robust learning outcomes in randomized learning frameworks.

\section{Conclusion}
\noindent In this paper, we employ the recursive calculation of the Moore-Penrose inverse to derive necessary and sufficient conditions for the training residuals to satisfy $\| e_L \| \leq \sqrt{r} \| e_{L-1} \|$ for each $L$ in SCN-III, the benchmark algorithm of SCN. This condition not only theoretically ensures convergence of training residuals as a geometric sequence but also enables rapid and accurate evaluations of random basis functions’ effectiveness in training error reductions, without necessitating the computation of the Moore-Penrose inverse of the output matrix of the hidden layer. Based on the conditions established, we propose the Recursive Moore-Penrose Inverse-SCN (RMPI-SCN). Compared to the classic SCN-III, the inequality constraint of RMPI-SCN consistently selects the most effective random candidate for the new basis function in each training iteration. This advancement not only ensures a faster convergence rate but also provides a more accurate assessment of the current uniform distribution’s capacity to yield suitable random basis functions. Simulation results across ten datasets confirm that RMPI-SCN outperforms SCN-III, exhibiting faster convergence rates and enhanced learning capabilities, particularly evident in the reduced training RMSE and higher correlation coefficients. Specifically, RMPI-SCN demonstrates significant improvements in function approximation tasks, achieving lower error rates on critical datasets.

\noindent Further research avenues regarding the supervisory mechanism of SCN are plentiful. Notably, the adaptive control of the learning rate $r$ significantly impacts SCN's performance, making a data-driven approach to controlling $r$ highly desirable. Moreover, the supervisory mechanism of SCN creates a subset in $L^2$ during each training iteration, where the basis functions within this subset contribute most effectively to reducing training errors. The mathematical exploration of this subset through probability measures or differential geometry could deepen our understanding of the learning and generalization capabilities of randomized learning models. Additionally, orthogonalizing the basis functions of SCN might yield interesting results regarding the model's interpretability, as it could facilitate clearer insights into how each function contributes to the overall prediction. Integrating optimization algorithms like PSO with the new inequality constraint may also significantly enhance SCN performance. Finally, this paper establishes a new inequality constraint for SCNs utilizing SLFNN. Future extensions of this constraint to other architectures, such as RSCNs, 2D-SCNs, and deep SCNs, present abundant opportunities for research.

\section*{Acknowledgement}
\noindent This work was supported by the National Key Research and Development Program of China (under Grant Number 2018AAA0100304). We would also like to express our gratitude to Pengxin Tian, who designed the training diagram presented in Figure 1.

\end{CJK}
\end{document}